\documentclass{article}

 \usepackage[main, final]{neurips_2026}

\usepackage{natbib}
\usepackage{enumitem}
\usepackage[utf8]{inputenc} 
\usepackage[T1]{fontenc}    
\usepackage{hyperref}
\hypersetup{
    colorlinks=true,
    linkcolor=blue,       
    citecolor=cyan,       
    urlcolor=blue         
}
\usepackage{url}            
\usepackage{booktabs}       
\usepackage{amsfonts}       
\usepackage{nicefrac}       
\usepackage{microtype}      
\usepackage{xcolor}         

\usepackage{graphicx}
\usepackage{wrapfig}
\usepackage{multirow}
\usepackage{makecell}
\usepackage{gensymb}
\usepackage{amsmath}
\usepackage{mathtools}
\usepackage{caption}
\usepackage{adjustbox}

\newtheorem{definition}{Definition}
\newtheorem{theorem}{Theorem}

\usepackage[table]{xcolor}
\usepackage{fontawesome5}

\title{Editing Large Language Models with Geometry-Aware Regularization}

\author{%
  Bingqing Liu$^{1}$\quad 
  Wei Liu$^{1}$ \quad Jun Wang$^{2}$ \quad
  Xiaobo Sun$^{3,\ast}$ \quad Yuhua Li$^{1,\ast}$ \quad Ruixuan Li$^{1,\ast}$ \\[0.5em]
  $^{1}$School of Computer Science and Technology, Huazhong University of Science and Technology\\
  $^{2}$iWudao Tech \quad
  $^{3}$Department of Human Genetics, School of Medicine, Emory University \\
  \texttt{\{liubingqing24, idc\_lw, idcliyuhua, rxli\}@hust.edu.cn}\\
  \texttt{jwang@iwudao.tech,xiaobo.sun@emory.edu}\\
  $^*$Corresponding authors.\quad \faGithub\ \href{https://github.com/lbq8942/GEdit}{Code}
}

\begin{document}

\maketitle

\begin{abstract}
Knowledge editing has become a key technique for updating knowledge in large language models. However, existing methods neglect the geometric structure of the knowledge representation space, leading to two key limitations: they cannot distinguish which pre-trained knowledge should be preferentially preserved (core vs. non-core) nor which new edits are inherently difficult (hard vs. easy). To address this, we propose \textsc{GEdit}, a \textbf{geometry-aware} framework built on a Low-Rank Plus Diagonal (LR+D) factor model that captures the intrinsic low-rank structure of Transformer FFN hidden representations. \textsc{GEdit} features two complementary mechanisms. First, the \textbf{low-rank regularization} applies targeted protection on a low-rank core knowledge subspace. Second, the \textbf{adaptive regularization} dynamically adjusts the regularization strength based on each edit request's editability: relaxing constraints for hard edits to ensure success, and tightening them for easy edits to better preserve pretrained knowledge. Extensive experiments on \texttt{GPT2-XL}, \texttt{GPT-J}, and \texttt{LLaMA-3-8B} under large-scale editing scenarios—including both mini-batch and fully sequential editing settings with up to \textbf{10,000 continuous updates}—show that \textsc{GEdit} consistently and significantly outperforms state-of-the-art baselines in both editing success and knowledge preservation. 

\end{abstract}

\section{Introduction}
\label{sec:introduction}

Large language models (LLMs) have become foundational repositories of world knowledge, encoding vast amounts of factual information within their parameters through extensive pretraining. However, this static knowledge base is inherently prone to obsolescence and inaccuracy: once training is complete, LLMs lack efficient mechanisms to incorporate new facts or rectify outdated information without prohibitively expensive full retraining. Knowledge editing has thus emerged as a critical research direction, aiming to enable precise, targeted modifications to a model's internal knowledge while preserving its general capabilities and unrelated facts \citep{mitchellfast,yao2023editing,wang2024knowledge,gu2024model,wang2025ace,zhang2025explainable,liumitigating}.

The central challenge in knowledge editing lies in balancing editing efficacy (successfully inserting new knowledge) with knowledge preservation (retaining unrelated pretrained knowledge). Despite recent progress, existing methods suffer from two critical limitations, both rooted in neglecting the underlying geometric structure of the knowledge representation space. First, they neglect the geometry of pretrained knowledge, applying uniform regularization across the entire space and overlooking the core knowledge subspace that deserves targeted protection \citep{fangalphaedit}. Second, they neglect the geometry of editing knowledge, failing to sense whether an edit will be easy or hard and treating all edits with identical regularization strength \citep{xutanglescore}. Consequently, this one-size-fits-all regularization is too strong for hard edits and too weak for easy edits. Together, these limitations cause existing methods to struggle in balancing edit success and knowledge preservation, often leading to rapid model collapse after just a few hundred updates \citep{thedewikibigedit}.

Motivated by these geometric shortcomings, we propose \textsc{GEdit}, a \textbf{geometry-aware} editing framework built on a Low-Rank Plus Diagonal (LR+D) factor model that captures the intrinsic low-rank structure of Transformer FFN hidden representations ~\citep{yu2023compressing,chenmixture}. \textsc{GEdit} addresses both limitations within a unified framework through two complementary strategies. First, \textbf{low-rank regularization} identifies a core knowledge subspace and strictly preserves it throughout editing, confining updates to the non-core knowledge subspace to protect pre-trained knowledge while also reducing time and space complexity. Second, \textbf{adaptive regularization} dynamically adjusts the regularization strength based on each edit request's editability: relaxing constraints for hard edits to ensure edit success, while tightening them for easy edits to enhance knowledge preservation.

We evaluate \textsc{GEdit} on two standard editing benchmarks, \textsc{CounterFact} and \textsc{ZsRE}, using three widely adopted LLMs of varying scales: \texttt{GPT2-XL} (1.5B), \texttt{GPT-J} (6B), and \texttt{LLaMA3} (8B). We utilize the widely used sequential editing protocol \citep{fangalphaedit} to evaluate its capability in maintaining both edit efficacy and locality over long-term updates. We employ argmax-based metrics \citep{dai2025eamet,liubetaedit} for strict evaluation and equip baselines with history-aware regularization \citep{fangalphaedit,fang2025hippocampal,guotowards,liubetaedit} for fair comparison under sequential editing. Under extensive testing—\textbf{10,000 sequential edits} across batch sizes of 1, 20, and 200, as well as varying edit orders—existing geometry-agnostic methods either collapse or suffer severe degradation in at least one setting, whereas \textsc{GEdit} consistently maintains high edit efficacy and locality across all configurations.  Furthermore, our \textbf{geometry-aware} framework is highly generalizable: both \textbf{low-rank and adaptive regularization} can be readily integrated into existing geometry-agnostic methods (e.g., MEMIT \citep{mengmass} and AlphaEdit \citep{fangalphaedit}), which substantially boosts their editing performance during long-term sequential editing.

\section{Related Work}
\label{sec:related_work}

Knowledge editing (KE) methods for large language models fall into two broad categories: \textit{parameter-preserving} and \textit{parameter-modifying}. 

\noindent\textbf{Parameter-Preserving Approaches.}
These methods keep pretrained weights intact. Memory-based retrieval approaches~\citep{dong2022calibrating, zheng2023can, hartvigsen2023aging, jiang2024learning, bi2025decoding,fei2026scaling} store edits in external memory and inject them during inference. Model augmentation methods~\citep{mitchell2022memory, huangtransformer, yu2024melo, wang2024wise} add trainable adapters into Transformer layers to encode new knowledge. Despite achieving high edit locality, these methods introduce additional inference cost and struggle to scale to large numbers of edits.

\noindent\textbf{Parameter-Modifying Approaches.}
These methods directly update model weights for permanent knowledge integration. Meta-learning based editors, including MEND~\citep{mitchellfast}, MALMEN~\citep{tan2024massive}, and RLEdit~\citep{li2025reinforced}, train hypernetworks to predict localized weight updates. A dominant line of work follows the \textit{locate-then-edit} paradigm. ROME~\citep{meng2022locating} performs rank-one updates for single edits, whereas MEMIT~\citep{mengmass}, EMMET~\citep{gupta2024unified}, and EAMET~\citep{dai2025eamet} extend this approach to support massive-scale batch editing. AlphaEdit~\citep{fangalphaedit} enforces null-space constraints to safeguard pretrained knowledge. Extending this approach, BetaEdit~\citep{liubetaedit} mitigates knowledge leakage caused by pseudo-null spaces and employs a dynamically evolving null space to jointly preserve both pretrained and previously edited knowledge.

Despite their success in single or batch editing settings, sequential or lifelong editing introduces fundamental challenges: catastrophic forgetting of earlier edits and  pretrained knowledge due to continuous parameter drift ~\citep{gupta2024model, li2024can, liunveiling, jiang2025neuron, hu2025knowledge, zhanguncovering, lirethinking}. To mitigate these issues, recent strategies fall into two categories. The first aims to reduce per-step weight perturbation, including sparsifying updates (RECT~\citep{gu2024model}, NMKE~\citep{liu2025edit}), clipping dominant singular values (PRUNE~\citep{maperturbation}), and preserving dominant singular directions (AdaEdit~\citep{li2025adaedit}, SPHERE~\citep{liu2026energyregularized}). The second focuses on preventing historical forgetting, typically by collecting covariance statistics of past edits~\citep{guotowards, fang2025hippocampal}.
\section{Preliminaries}
\label{sec:prelim}

\paragraph{Problem Formulation}
A single edit is defined as a quadruple $\mathcal{E} = (s, r, o_{\text{ori}}, o_{\text{tar}})$, where $s$ is the subject entity (e.g., United Kingdom), $r$ the relation (e.g., The Prime Minister of), $o_{\text{ori}}$ the original object (e.g., Boris Johnson), and $o_{\text{tar}}$ the target object (e.g., Keir Starmer). Given a query prompt $x$ constructed from $s$ and $r$ (e.g., ``The Prime Minister of the United Kingdom is''), the goal of knowledge editing is to update the LLM's parameters so that it outputs $o_{\text{tar}}$ instead of $o_{\text{ori}}$.

\paragraph{Locate-then-Edit Framework}
We adopt the \textit{locate-then-edit} paradigm \citep{meng2022locating,mengmass}, which first identifies causal components for a fact and then applies a minimal parameter update.

\noindent\textit{Where is knowledge stored?}
Factual knowledge in LLMs is primarily encoded in feed-forward network (FFN) layers \citep{geva2021transformer,meng2022locating}. Each FFN layer acts as a key-value memory: the second linear layer $\mathbf{W}$ maps a key $\mathbf{k}$ to a value $\mathbf{v} = \mathbf{W}\mathbf{k}$. Under the \textit{locate-then-edit} (LTE) framework, a knowledge triple $(s, r, o)$ is stored as a key-value pair $(\mathbf{k}, \mathbf{v})$, where $\mathbf{k}$ is the hidden activation of subject $s$, and $\mathbf{v}$ encodes the attribute $o$ under relation $r$.

\noindent\textit{How is knowledge updated?}
An edit $\mathcal{E} = (s, r, o_{\text{ori}}, o_{\text{tar}})$ involves two pairs: $(\mathbf{k}, \mathbf{v}_{\text{ori}})$ and $(\mathbf{k}, \mathbf{v}_{\text{tar}})$. Since $\mathbf{W}\mathbf{k} = \mathbf{v}_{\text{ori}}$, we seek a perturbation $\Delta$ with $(\mathbf{W} + \Delta)\mathbf{k} = \mathbf{v}_{\text{tar}}$ while preserving unrelated knowledge. Formally, let $\mathbf{K}_1 \in \mathbb{R}^{d \times n_e}$, $\mathbf{V}_1 \in \mathbb{R}^{d_{\text{ffn}} \times n_e}$ collect the keys and target values of $n_e$ edits, and $\mathbf{K}_0 \in \mathbb{R}^{d \times n_p}$, $\mathbf{V}_0 \in \mathbb{R}^{d_{\text{ffn}} \times n_p}$ those of $n_p$ pretrained facts to preserve. The optimal weight update $\Delta \in \mathbb{R}^{d_{\text{ffn}} \times d}$ is then obtained by solving the following constrained least-squares problem \citep{mengmass} with $\beta > 0$ balancing edit success and knowledge preservation:
\begin{equation}
\label{eq:memit_rule}
\Delta = \arg\min_{\tilde{\Delta}} 
\|(\mathbf{W} + \tilde{\Delta})\mathbf{K}_1 - \mathbf{V}_1\|_F^2 
+ \beta \|(\mathbf{W} + \tilde{\Delta})\mathbf{K}_0 - \mathbf{V}_0\|_F^2,
\end{equation}

\section{Method}
\label{sec:method}

We first analyze the low-rank structure of FFN hidden representations. Based on this analysis, we introduce a probabilistic model—the Low-Rank Plus Diagonal (LR+D) factor model—to capture this low-rank geometry and enable low-rank regularization. Then, for a given edit, we define its editability and propose an adaptive regularization strategy.

\subsection{Low-Rank Structure of FFN Hidden Representations}
Empirical studies have demonstrated that hidden representations within the FFN layers of Transformers tend to reside in low-dimensional subspaces~\citep{aghajanyan2021intrinsic,yu2023compressing,chenmixture}. To investigate this geometric property, we collect FFN hidden representations from a large-scale Wikipedia corpus, following the protocol of \citet{meng2022locating}. Specifically, we sample 100,000 sentences and extract FFN hidden vectors from layer~8 of two LLMs: \texttt{LLaMA3-8B} ($d = 14,\!336$) and \texttt{GPT-J-6B} ($d = 16,\!384$).\footnote{Layer~8 is commonly used for knowledge editing in prior work~\citep{fangalphaedit,lirethinking}.} For each LLM, we stack its FFN hidden representations column-wise to form a separate  matrix $\mathbf{K}_0 \in \mathbb{R}^{d \times N}$, where $N \approx 4 \times 10^7$.

\begin{wrapfigure}{r}{0.45\linewidth}
  \centering
  \includegraphics[width=0.95\linewidth]{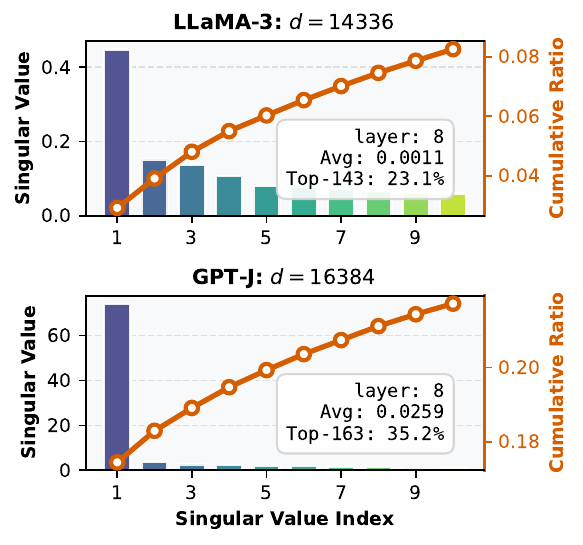}
  \caption{Top-10 singular values (left axis) and their cumulative ratio of the total singular value sum (right axis).}
  \label{fig:lowrank}
\end{wrapfigure}

We perform singular value decomposition (SVD) on each $\mathbf{K}_0$ and present the resulting singular value spectra in Figure~\ref{fig:lowrank}. Both spectra exhibit a steep long-tailed decay:  the average singular value approaches zero, and the top 1\% of singular values account for over 23\% of the total sum. This behavior reveals the latent low-rank structure of the FFN hidden representations. 

To interpret this low-rank structure from a knowledge representation perspective, let $\mathbf{U} \in \mathbb{R}^{d \times r}$ be the top-$r$ principal subspace that captures most factual knowledge, and $\mathbf{M} = \mathbf{U}^\perp$ its orthogonal complement, which encodes limited factual information. Modifying $\mathbf{U}$ risks disrupting much pretrained knowledge, whereas updates confined to $\mathbf{M}$ minimize disruption. This insight motivates our low-rank regularization, which strictly protects the core knowledge subspace  $\mathbf{U}$.

\subsection{Geometry-Aware Low-Rank Regularization}

In the standard \emph{locate-then-edit} framework (Eq.~\ref{eq:memit_rule}), the regularization term $\|(\mathbf{W}+\Delta) \mathbf{K}_0-\mathbf{V}_0\|_F^2$ is \textit{geometry-agnostic} for pretrained knowledge preservation. To address this issue, we introduce a $\mathbf{U}$-subspace-aware regularizer derived from a probabilistic model of FFN hidden representations.

\paragraph{Low-Rank Plus Diagonal (LR+D) Factor Model}
Let $\mathbf{k} \in \mathbb{R}^d$ be a hidden vector sampled from the model’s FFN hidden space. Under the Low-Rank Plus Diagonal (LR+D) factor model~\citep{Fan2011HIGHDC,zhao2016low,liutkus2017diagonal}, $\mathbf{k}$ is assumed to follow the generative process:
\begin{equation}
\mathbf{k} = \mathbf{b}+\mathbf{U} \mathbf{z} + \boldsymbol{\varepsilon},
\label{eq:factor_model}
\end{equation}
where $\mathbf{b} = \mathbb{E}[\mathbf{k}] \in \mathbb{R}^d$ is the mean vector, $\mathbf{z} \sim \mathcal{N}(\mathbf{0}, \mathbf{I})$ represents a latent semantic code, $\mathbf{U} \in \mathbb{R}^{d \times r}$ spans the dominant semantic subspace (with $r \le d$), $\boldsymbol{\varepsilon} \sim \mathcal{N}(\mathbf{0}, \mathbf{D})$ captures dimension-wise residual variation, and $\mathbf{z}$ and $\boldsymbol{\varepsilon}$ are independent. Here $\mathbf{D} = \mathrm{diag}(h_1, \dots, h_d)$ is a diagonal covariance matrix, where each $h_i > 0$ denotes the residual variance along the $i$-th dimension. This model generalizes several important cases; for example, when $\mathbf{U} = \mathbf{0}$, it reduces to an independent-component Gaussian assumption over the FFN hidden representations.

\paragraph{Geometry-Aware Regularization}
Under the probabilistic model of the key vector $\mathbf{k}$ introduced above, we can quantify how a weight update $\Delta$ perturbs the corresponding value vector $\mathbf{v}$. Recall that $\mathbf{v}$ is not an independent variable; it is determined by $\mathbf{k}$ and the weight matrix $\mathbf{W}$ via $\mathbf{v} = \mathbf{W}\mathbf{k}$.

Taking the expectation over the distribution of $\mathbf{k}$, the expected output perturbation induced by $\Delta$ is:
\begin{equation}
\mathbb{E}_{\mathbf{k}} \big[ \|(\mathbf{W}+\Delta) \mathbf{k}-\mathbf{v}\|_2^2 \big] 
= \mathrm{Tr}\big( \Delta^\top \Delta \mathbf{U}\mathbf{U}^\top \big) + \mathrm{Tr}\big( \Delta^\top \Delta \bar{\mathbf{D}} \big),
\label{eq:exp_dk}
\end{equation}
where $\bar{\mathbf{D}}=\mathbf{b}\mathbf{b}^\top+\mathbf{D}$ and $\mathrm{Tr}$ denotes the trace. (See Appendix~\ref{subapp:property} for the derivation.)

To preserve pretrained knowledge, we minimize this expected output perturbation. Combining this regularization term with the edit efficacy objective yields our final knowledge editing loss:\footnote{Without loss of generality, we adopt the Frobenius norm for the edit efficacy term \citep{mengmass,fang2025hippocampal}, though an equality constraint could alternatively be used \citep{meng2022locating,gupta2024unified} (see Appendix \ref{subapp:low_adapt}).}
\begin{equation}
\min_{\Delta} \; \left\| (\mathbf{W} + \Delta) \mathbf{K}_1 - \mathbf{V}_1 \right\|_F^2
+ \beta \cdot \mathrm{Tr}\big( \Delta^\top \Delta \mathbf{U}\mathbf{U}^\top \big)
+ \mu \cdot \mathrm{Tr}\big( \Delta^\top \Delta \bar{\mathbf{D}} \big),
\label{eq:lr_obj}
\end{equation}
where $\beta>0$ and $\mu > 0$ balance edit efficacy against knowledge preservation. This $\mathbf{U}$-subspace-aware regularizer penalizes weight updates that align with the core knowledge subspace $\mathbf{U}$, thereby protecting critical knowledge from being overwritten.

\paragraph{Closed-Form Update}
Define the residual $\mathbf{R}_1 = \mathbf{V}_1 - \mathbf{W} \mathbf{K}_1$. The optimization problem in Eq.~\ref{eq:lr_obj} admits the following closed-form solution (see Appendix~\ref{subapp:closed_form} for the derivation):
\begin{equation}
\label{eq:update_combined}
\Delta = \mathbf{R}_1 \mathbf{K}_1^\top \big( \mathbf{K}_1 \mathbf{K}_1^\top + \beta \mathbf{U}\mathbf{U}^\top + \mu \bar{\mathbf{D}} \big)^{-1}.
\end{equation}

To protect the core knowledge subspace $\mathbf{U}$, we consider the limit $\beta \to \infty$, which suppresses updates along directions spanned by $\mathbf{U}$. The update then reduces to (see Appendix~\ref{app:alphaedit_conn} for the derivation):
\begin{equation}
\label{eq:gedit_null_limit}
\boxed{\Delta_{\mathbf{G}} = \lim_{\beta \to \infty} \Delta 
= \mathbf{R}_1 \mathbf{K}_1^\top \mathbf{M} \big( \mathbf{M}^{\top}(\mathbf{K}_1 \mathbf{K}_1^\top + \mu \bar{\mathbf{D}})\mathbf{M} \big)^{-1} \mathbf{M}^{\top}},
\end{equation}
where $\mathbf{M}\in \mathbb{R}^{d\times (d-r)}$ is a matrix whose columns form a basis for the orthogonal complement of $\mathbf{U}$ (i.e., $\mathbf{M}^{\top}\mathbf{U} = \mathbf{0}$). This limiting update has two features. First, it only affects knowledge residing in the non-core subspace $\mathbf{M}$. Second, its inversion has time complexity $O((d - r)^3)$ and space complexity $O((d - r)^2)$, compared to $O(d^3)$ and $O(d^2)$ for standard LTE methods—an advantage that becomes significant when working with large $d$ and editing multiple layers.
 
\paragraph{Estimation of $\mathbf{U}$, $\mathbf{D}$, and $\mathbf{b}$}
Under the LR+D model, the covariance of $\mathbf{k}$ satisfies
$\mathrm{Cov}(\mathbf{k}) = \mathbf{U}\mathbf{U}^\top + \mathbf{D}$
(see Appendix~\ref{subapp:property} for the derivation).
This suggests that $\mathbf{U}$ and $\mathbf{D}$ can be estimated from the empirical covariance of FFN hidden representations, while $\mathbf{b}$ is estimated from the sample mean.
To this end, we collect $N = 10^7$ hidden vectors $\{\mathbf{k}_i\}_{i=1}^N$ by passing Wikipedia text \citep{mengmass,meng2022locating} through the model, and compute the sample mean $\bar{\mathbf{k}} = \frac{1}{N} \sum_{i=1}^N \mathbf{k}_i$ and the empirical covariance matrix
$\mathbf{S}_0 = \frac{1}{N} \sum_{i=1}^N (\mathbf{k}_i - \bar{\mathbf{k}})(\mathbf{k}_i - \bar{\mathbf{k}})^\top$.
We then perform an eigendecomposition
$\mathbf{S}_0 = \mathbf{Q} \boldsymbol{\Lambda} \mathbf{Q}^\top$,
with eigenvalues $\lambda_1 \geq \lambda_2 \geq \cdots \geq \lambda_d \geq 0$ arranged in descending order along the diagonal of $\boldsymbol{\Lambda}$, and $\mathbf{Q} = [\mathbf{q}_1, \dots, \mathbf{q}_d]$ as the corresponding eigenvectors. We estimate $\mathbf{U}$, $\mathbf{D}$, and $\mathbf{b}$ as follows:
\begin{align}
\mathbf{U} &= [\mathbf{q}_1,\cdots,\mathbf{q}_r] \, \boldsymbol{\Lambda}_{1:r,1:r}^{1/2}, \qquad
\mathbf{D}= \mathrm{diag}\big( \mathbf{S}_0 - \mathbf{U}\mathbf{U}^\top \big), \qquad
\mathbf{b} = \bar{\mathbf{k}}.
\end{align}
This construction is theoretically grounded in the Eckart--Young--Mirsky theorem~\citep{golub2013matrix}: under the rank-$r$ constraint, $\mathbf{U}$ captures the most informative subspace of $\mathrm{Cov}(\mathbf{k})$, and $\mathbf{D}$ retains the remaining diagonal residual.

The next critical step is to select an appropriate rank $r$. Intuitively, models with a larger FFN hidden dimension $d$ tend to require a larger $r$, whereas those exhibiting stronger low-rank structure admit a smaller $r$. To automatically capture these opposing trends, we adopt an energy-based criterion for adaptive rank selection:
\begin{equation}
\label{eq:gamma}
r = \min\left\{ j : \frac{\sum_{i=1}^j \lambda_i}{\sum_{i=1}^{d} \lambda_i} \geq \gamma \right\},
\end{equation}
where $\gamma \in (0,1]$ is a prescribed energy retention threshold (e.g., $\gamma = 0.85$). This rule selects the smallest number of principal components sufficient to capture at least a $\gamma$-fraction of the total spectral energy, thereby directly reflecting the empirical low-rank structure of the FFN representation space and adapting naturally to the FFN hidden dimension $d$.

\subsection{Geometry-Aware Adaptive Regularization}
Existing approaches neglect the geometric structure of edit requests and apply a uniform regularization strength to all of them. Ideally, a hard edit requires relaxed regularization (smaller \(\mu\)) to ensure successful editing, whereas an easy edit benefits from stronger regularization (larger \(\mu\)) to better preserve pretrained knowledge. We therefore propose an adaptive mechanism that selects \(\mu\) on the fly based on the hardness of each edit.

To quantify edit hardness, we introduce the notion of \textit{editability}. This notion is grounded in the following guiding principle, which links geometry to edit difficulty.
\begin{quote}
    \textit{Edits aligned with knowledge-sparse subspaces are more ``editable,'' whereas those lying in knowledge-dense subspaces are less so since they likely conflict with more pretrained facts, inducing larger knowledge preservation loss (Eq. \ref{eq:memit_rule}).}
\end{quote}

To formalize this principle, we have the following definition of \textit{editability}.
\begin{definition}[Editability]
For an edit vector $\mathbf{k}_e \in \mathbb{R}^d$, we define its editability as
\begin{equation}
\mathcal{E}(\mathbf{k}_e)
= \frac{1}{\mu}\sum_{j=1}^d
\sigma_j(\mathbf{k}_e^\top \mathbf{q}_j)^2,
\label{eq:editability}
\end{equation}
where $\{\mathbf{q}_j\}_{j=1}^d$ are the eigenvectors of the second-moment matrix $\mathbf{C}_0=\mathbf{K}_0\mathbf{K}_0^\top$, and $\{\sigma_j\}_{j=1}^d$ are positive weights satisfying $0<\sigma_1\leq \sigma_2\leq\cdots\leq\sigma_d$. Larger weights are assigned to knowledge-sparse directions and smaller weights to knowledge-dense directions. The regularization strength $\mu>0$ inversely scales editability.
\end{definition}
\begin{table}[htbp]
\centering
\small
\caption{Unification of output perturbation via editability.}
\label{tab:method_comparison}
\begin{tabular}{lccl}
\toprule
\textbf{Method} & \textbf{Weight Update} $\Delta$ & \textbf{Editability} $\mathcal{E}(\mathbf{k}_e)$& \textbf{Output Perturbation} \\
\midrule
MEMIT & $\mathbf{r}_e \mathbf{k}_e^\top \left( \mu \mathbf{C}_0 + \mathbf{k}_e \mathbf{k}_e^\top \right)^{-1}$ & $\frac{1}{\mu} \mathbf{k}_e^\top \mathbf{C}_0^{-1} \mathbf{k}_e$ & \multirow{3}[2]{*}{$\Delta \mathbf{k}_e = \dfrac{\mathcal{E}(\mathbf{k}_e)}{1 + \mathcal{E}(\mathbf{k}_e)} \mathbf{r}_e$} \\
\cmidrule{1-3}
ROME & $\mathbf{r}_e \left( \mathbf{k}_e^\top (\mu \mathbf{C}_0)^{-1} \mathbf{k}_e + a \mathbf{I} \right)^{-1} \mathbf{k}_e^\top (\mu \mathbf{C}_0)^{-1}$ & $\frac{1}{\mu a} \mathbf{k}_e^\top \mathbf{C}_0^{-1} \mathbf{k}_e$  & \\
\cmidrule{1-3}
AlphaEdit & $\mathbf{r}_e \mathbf{k}_e^\top \mathbf{P} \left( \mathbf{k}_e \mathbf{k}_e^\top \mathbf{P} + \mu \mathbf{I} \right)^{-1}$ & $\frac{1}{\mu} \mathbf{k}_e^\top \mathbf{P} \mathbf{k}_e$  & \\
\bottomrule
\end{tabular}
\end{table}
\begin{wrapfigure}[18]{r}{0.45\linewidth}
  \centering
  \includegraphics[width=1.0\linewidth]{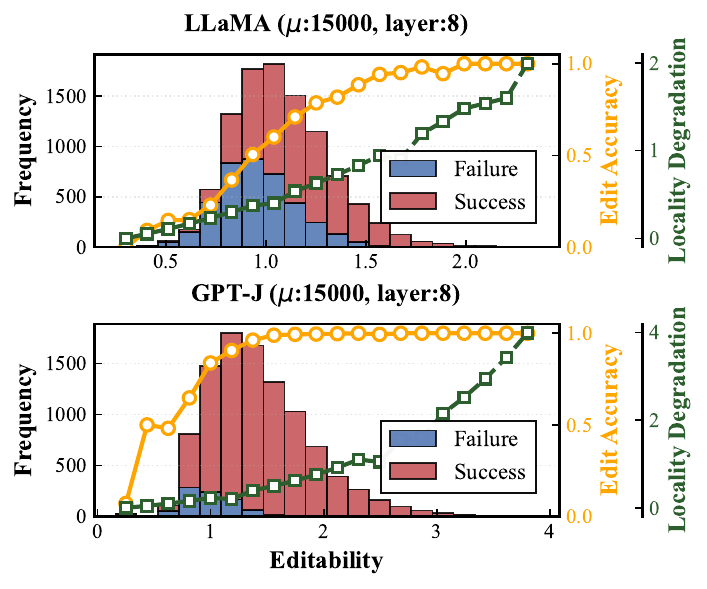}
  \caption{Distribution of editability across 10,000 edits from the CounterFact dataset, along with the editing success rate and locality degradation within each editability bin.}
  \label{fig:ada_mu}
\end{wrapfigure}
A higher $\mathcal{E}(\mathbf{k}_e)$ indicates that the edit aligns more strongly with knowledge-sparse directions and is subject to weaker regularization (smaller $\mu$). In our implementation, we set $\sigma_j = 1/\lambda_j$, where $\lambda_j$ is the eigenvalue of $\mathbf{C}_0$ associated with $\mathbf{q}_j$. In this case the editability takes the form of the squared Mahalanobis distance (see Appendix \ref{subapp:mahal_derivation} for details):
\begin{equation}
\mathcal{E}(\mathbf{k}_e) =\frac{1}{\mu} \mathbf{k}_e^\top \mathbf{C}_0^{-1} \mathbf{k}_e.
\label{eq:editability_c}
\end{equation}

Under the LR+D model, when edits are constrained to the non-core knowledge subspace $\mathbf{M}$, we correspondingly define the editability restricted to $\mathbf{M}$ as
\begin{equation}
\mathcal{E}(\mathbf{k}_e) =\frac{1}{\mu} \mathbf{k}_e^\top \mathbf{M} (\mathbf{M}^\top \bar{\mathbf{D}} \mathbf{M})^{-1} \mathbf{M}^\top \mathbf{k}_e,
\label{eq:editability_d}
\end{equation}
where the edit vector and the inverse are restricted to the non-core subspace (see Appendix~\ref{subapp:editability_derivation} for details).

We now formally establish the relationship between editability and edit success: higher editability leads to greater edit success, as stated in the theorem below.

\begin{theorem}
\label{th:ee}
Let $(\mathbf{k}_e, \mathbf{v}_e)$ be a knowledge pair to be edited, and let $\mathbf{r}_e = \mathbf{v}_e - \mathbf{W}\mathbf{k}_e$ denote the corresponding target output residual. Let $\mathcal{E}(\mathbf{k}_e)$ be the editability. Then the output perturbation $\Delta_{G}\mathbf{k}_e$ induced by \textsc{GEdit} satisfies
\[
    \Delta_{G}\mathbf{k}_e = \frac{\mathcal{E}(\mathbf{k}_e)}{1 + \mathcal{E}(\mathbf{k}_e)} \, \mathbf{r}_e.
\]
\end{theorem}

The proof is provided in Appendix~\ref{subapp:theorem_editability}. As $\mathcal{E}(\mathbf{k}_e)$ increases, the scaling factor $\mathcal{E}(\mathbf{k}_e)/(1 + \mathcal{E}(\mathbf{k}_e))$ monotonically approaches $1$, implying that the output perturbation $\Delta_{G}\mathbf{k}_e$ becomes increasingly aligned with the target residual $\mathbf{r}_e$, thereby improving editing accuracy. In the limit $\mathcal{E}(\mathbf{k}_e) \to \infty$, the perturbation exactly recovers $\mathbf{r}_e$, guaranteeing perfect editing success.

Since different edits have varying degrees of editability, their editing accuracy and locality also differ accordingly. To illustrate this phenomenon, we conduct experiments on 10,000 edits from the CounterFact dataset using \texttt{LLaMA} and \texttt{GPT-J}, measuring their editability (Eq. \ref{eq:editability_d} with $\mu=15000$) alongside editing success and locality degradation. An edit is considered successful if the model generates the edit output $o_{\text{tar}}$. Locality degradation is quantified as the change in logits for the locality output $o_{\text{ori}}$ before and after editing. As shown in Figure~\ref{fig:ada_mu}, two key observations emerge:
First, edits with low editability tend to fail but cause minimal locality degradation, whereas highly editable edits are more likely to succeed yet induce significant locality degradation. Second, using fixed regularization is suboptimal: with $\mu$ fixed at 15,000, the value is too large for \texttt{LLaMA}—leading to excessive editing failures—and too small for \texttt{GPT-J}, resulting in severe locality degradation. 

Both theoretical analysis and empirical results indicate that editing success and locality degradation can be effectively balanced by modulating editability—motivating the use of editability-based regularization.  Specifically, by setting $\mathcal{E}(\mathbf{k}_e) = c$, where \( c > 0 \) is a hyperparameter representing the desired editability (e.g., \( c = 2 \)), we obtain an edit-adaptive regularization strength
$\mu = \frac{1}{c} \mathbf{k}_e^\top \mathbf{M} (\mathbf{M}^\top \mathbf{D} \mathbf{M})^{-1} \mathbf{M}^\top \mathbf{k}_e$. We note that Theorem~\ref{th:ee} is \textbf{agnostic} to dataset, model architecture, and even editing method, thereby providing a unified framework for adapting regularization strength—specifically, by setting the editability to $c$. Table~\ref{tab:method_comparison} summarizes the application of our framework to several popular editors; see Appendix~\ref{subapp:ada_adapt} for details.  Interestingly, Table~\ref{tab:method_comparison} shows that different methods have different definitions of editability, which fundamentally explains why they have different settings of the regularization strength. For example, $\mu$ is set to 10 in AlphaEdit's paper \citep{fangalphaedit}, while it is set to 15,000 for MEMIT \citep{mengmass}. Our adaptive regularization framework can also be extended to sequential editing by jointly preserving pretrained and previously edited knowledge \citep{fangalphaedit,fang2025hippocampal,guotowards,liubetaedit} (see Appendix~\ref{subapp:ada_adapt} for details).

\section{Experiments}
\subsection{Experimental Setup}
\label{sec:setup}

\paragraph{LLMs, Datasets and Evaluation Protocol} We edit three widely used LLMs—\texttt{GPT2-XL}, \texttt{GPT-J}, and \texttt{LLaMA3-8B-Instruct}—on two standard factual editing benchmarks: \textsc{CounterFact} \citep{meng2022locating} and \textsc{ZsRE} \citep{levy2017zero}. Following standard practice \citep{fangalphaedit,lirethinking}, we report three metrics: \textit{Efficacy (Eff.)}, \textit{Generality (Gen.)}, and \textit{Locality (Loc.)}. Specifically, \textit{Eff.} and \textit{Gen.} measure whether the model can correctly predict the target $o_{tar}$ given the original and paraphrased edit prompts, respectively, while \textit{Loc.} evaluates the model's prediction accuracy on irrelevant knowledge. A good editor should maintain close \textit{Loc.} performance before and after editing. Prior work often overestimates editing performance by using probability-comparison-based calculations for these metrics \citep{gupta2024unified,fangalphaedit}. To address this, we adopt a strict argmax-based evaluation \citep{dai2025eamet,liu2026energyregularized}. Dataset examples and detailed metric calculations are provided in Appendix \ref{app:data_metric}. Another often-overlooked factor during evaluation is the batch size and edit order, which can substantially affect editing performance. Existing methods use inconsistent batch sizes (e.g., 1 \citep{guotowards}, 20 \citep{li2025reinforced}, and beyond \citep{fangalphaedit,liu2025edit,liu2026energyregularized,parksuit}) and rely on a single fixed edit order \citep{fangalphaedit,liubetaedit}. To ensure a more comprehensive evaluation, we perform 10,000 sequential edits under three batch sizes: 1, 20, and 200, and average the results over 3 different edit orders.

\begin{figure}[t]
  \centering
  \includegraphics[width=1.0\linewidth]{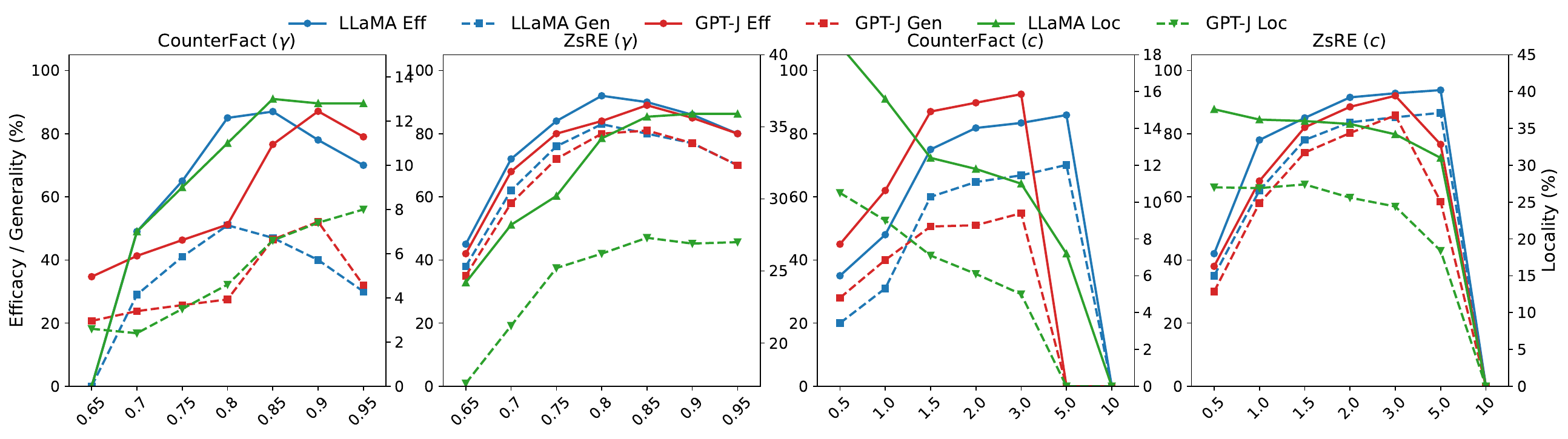}
\caption{
Effect of hyperparameters $\gamma$ (energy threshold) and $c$ (desired editability) on editing performance. Metrics on the left axis: efficacy and generality; metric on the right axis: locality.
}
  \label{fig:hp}
\end{figure}

\paragraph{Compared Baselines} We compare against a wide range of baselines: LoRA-based FT \citep{hu2022lora}; hypernetwork-based RLEdit \citep{li2025reinforced}; locate-then-edit (LTE) methods (\textsc{EMMET} \citep{gupta2024unified}, \textsc{AlphaEdit} \citep{fangalphaedit}, \textsc{HSE} \citep{fang2025hippocampal}); and regularized sequential editing methods (\textsc{RECT} \citep{gu2024model}, \textsc{PRUNE} \citep{maperturbation}, \textsc{AdaEdit} \citep{li2025adaedit}, \textsc{SimIE} \citep{guotowards}, \textsc{SPHERE} \citep{liu2026energyregularized}). Following \citet{fangalphaedit}, for all LTE methods, we edit layers \{13,14,15,16,17\} in \texttt{GPT2-XL}, layers \{3,4,5,6,7,8\} in \texttt{GPT-J}, and layers \{4,5,6,7,8\} in \texttt{LLaMA3}. All LTE methods can be readily combined with the edit protection mechanism from \citep{fangalphaedit}, which significantly improves their sequential editing performance \citep{guotowards,fang2025hippocampal,liubetaedit}. To ensure a fair comparison—often overlooked in prior work \citep{fang2025hippocampal,liu2025edit,liu2026energyregularized}—we integrate this mechanism into all LTE methods (see Appendix \ref{subapp:adapt} for details). For the five regularized sequential editing methods, we adopt \textsc{HSE} as the base editor.

For \textsc{GEdit}, we investigate two key hyperparameters. $\gamma$ sets the energy retention ratio that identifies the core knowledge subspace, while $c$ determines the desired editability and adaptively adjusts the regularization strength $\mu$. A higher $\gamma$ protects a larger core knowledge subspace (the mapping between $\gamma$ and the resulting rank $r$ is provided in Appendix~\ref{subapp:low_adapt}), whereas a lower $c$ imposes stronger regularization. Their effects are illustrated in Figure~\ref{fig:hp}. Performance is more sensitive to $\gamma$; for example, on GPT-J/CounterFact, $\gamma=0.9$ outperforms $\gamma=0.85$ and $0.95$ by $\approx$10 points in \textit{Eff.}

In our experiments, we aim to answer the following research questions:
\begin{itemize}[nosep, topsep=1pt, leftmargin=1.5em]
    \item \textbf{RQ1:} What is the editing performance of \textsc{GEdit}?
    \item \textbf{RQ2:} What is the impact of low-rank regularization?
    \item \textbf{RQ3:} What is the impact of adaptive regularization?
\end{itemize}

\begin{table*}[h]
\centering
\caption{
    Editing performance over 10,000 sequential edits under varying batch sizes (1, 20 and 200).
    Best and second-best results are \textbf{bolded} and \underline{underlined}, respectively. 
}
\label{tab:results}
\setlength{\tabcolsep}{2.5pt}
\resizebox{\textwidth}{!}
{
\begin{tabular}{l l c c c c c c cc c c c cc c c c c c}
\toprule
\multirow{3}{*}{\textbf{Method}} & 
& \multicolumn{9}{c}{\textbf{CounterFact}} 
& \multicolumn{9}{c}{\textbf{ZsRE}} \\
\cmidrule(lr){3-11} \cmidrule(lr){12-20}
&&\multicolumn{3}{c}{\textbf{bs = 1}}&\multicolumn{3}{c}{\textbf{bs = 20}}&\multicolumn{3}{c}{\textbf{bs = 200}}&\multicolumn{3}{c}{\textbf{bs = 1}}&\multicolumn{3}{c}{\textbf{bs = 20}}&\multicolumn{3}{c}{\textbf{bs = 200}}\\
\cmidrule(lr){3-5} \cmidrule(lr){6-8}  \cmidrule(lr){9-11} \cmidrule(lr){12-14} \cmidrule(lr){15-17} \cmidrule(lr){18-20} 
& 
& \textbf{Eff.}& \textbf{Gen.} & \textbf{Loc.}
& \textbf{Eff.} & \textbf{Gen.} & \textbf{Loc.}& \textbf{Eff.} & \textbf{Gen.} & \textbf{Loc.} & \textbf{Eff.} & \textbf{Gen.} & \textbf{Loc.} & \textbf{Eff.} & \textbf{Gen.} & \textbf{Loc.} & \textbf{Eff.} & \textbf{Gen.} & \textbf{Loc.}  \\
\midrule
\midrule
Pre-edited && 0.40&0.50&10.6&0.40&0.50&10.6 &0.40&0.50&10.6&0.00 & 0.00 & 24.3 &0.00 & 0.00 & 24.3& 0.00 & 0.00 & 24.3 \\
\midrule
FT& \multirow{12}{*}{\rotatebox{90}{GPT2-XL-1.5B}} 
& 0.00 &  0.00 &  0.00 &0.00 & 0.00 &0.00 & 7.10 & 2.50 & 0.10
& 0.00 &  0.00 &  0.00  & 0.00 & 0.00 & 0.00 &10.3 & 3.60 & 4.20 \\
RLEdit&  
& 0.00 &  0.00 &  0.00 &33.5 & 21.8 & 2.20 &  45.3 & 25.5 & 2.80 
& 0.00 &  0.00 &  0.00  & 48.9 & 32.3 &16.8 & 37.3 & 11.1 & 11.2  \\
RECT&   
&23.6 &  13.1 & 4.10 &54.5 & \underline{33.6} & 6.40  &69.5 & 37.9 & 6.90 
& 64.0 & 50.8 & 24.1 & 67.9 & \underline{56.2} & 24.1 & 66.6 & 53.5 & 24.5  \\
PRUNE&  
&20.2 &  10.3 & 4.00&58.7 & 30.7 & 5.60  &68.7 & 38.7 & 6.50 
& 63.0 & 51.8 & 24.3& \underline{66.5} & 52.4 & 24.2 & 65.0 & 53.2 & 24.2  \\
AdaEdit&  
& 20.4 &  11.0 & 4.30 & 53.8 & 30.5 & \underline{6.80}& 63.8 & 37.8 & 6.90 
& 61.4 & 49.3 & 24.2 & 64.0 & 53.5 & 24.3 & 63.0 & 51.5 & 24.3 \\
SPHERE&  
& 31.2 &  20.1 & \underline{5.00}& 50.4 & 27.9 &7.10  & 61.4 & 32.9 & 7.20 
& 57.0 & 46.7 & \textbf{24.6} & 60.2 & 50.4 & \textbf{25.0}& 61.4 & 49.2 & 24.7 \\
SimIE&  
&27.1 &  18.2 & 4.20 & 47.9 & 24.4 & \textbf{7.90} &  55.9 & 27.4 & \underline{7.30} 
&53.3 & 38.3 & \underline{24.3} & 51.9 & 39.9 & 24.5& 58.8 & 34.7 & \textbf{25.0}  \\
EMMET&  
& 23.7 &  11.2 &  3.40&  \underline{60.5} & 32.7 & 3.80  & \textbf{75.5} & \textbf{43.2} & 4.10 
& 50.3 & 32.3 & 11.5 & 54.8 & 34.7 & 15.6 & 53.3 & 32.3 & 18.9 \\
HSE&   
&18.2 &  9.10 & 3.80& 59.3 & 28.6 & 5.10 & 70.5 & 39.7 & 6.40 
& \underline{67.3} & 52.3 & 24.0 & 66.2 & 53.7 & 24.2 &\underline{68.0} & \underline{54.9} & \underline{24.7} \\
AlphaEdit& 
& \underline{67.9} &  \underline{32.3} & 2.80 & 55.6 & 31.8 & 2.10 & 61.8 & 25.6 & 2.40 
& 64.3 & \underline{56.0} & 18.7 & 55.2 & 41.5 & 19.3 & 35.4 & 25.8 & 14.3 \\
\textbf{\textsc{GEdit}}& 
& \textbf{76.4} & \textbf{40.1} & \textbf{5.20} & \textbf{72.1} & \textbf{40.8} & 6.40 & \underline{71.5} & \underline{42.6} & \textbf{7.40} 
& \textbf{73.1} & \textbf{61.3} &24.0& \textbf{71.1} & \textbf{59.6} & \underline{24.7} & \textbf{73.9} & \textbf{57.1} & 24.1 \\
\midrule
Pre-edited && 0.40&0.50&14.3&0.40&0.50&14.3 &0.40&0.50&14.3&0.00 & 0.00 & 27.0 &0.00 & 0.00 & 27.0& 0.00 & 0.00 & 27.0 \\
\midrule
FT& \multirow{11}{*}{\rotatebox{90}{GPT-J-6B}} 
& 0.00 &  0.00 &  0.00 & 0.00 & 0.00 & 0.00 & 5.30 & 2.10 & 0.50 
& 0.00 &  0.00 &  0.00  & 0.00 & 0.00 & 0.00 & 0.00 & 0.00 & 0.00 \\
RLEdit&  
& 0.00&0.00&0.00 & 35.6 & 17.7 & 1.20 & 65.6 & 37.3 & 4.50 
& 0.00 & 0.00 & 0.00 & 70.5 & 53.7 & 10.2 & 75.6 & 61.4 & 18.4  \\
RECT&   
& 30.2 & 17.5 & 2.60 &  80.2 & 68.1 & 6.90  & 91.2 & 60.5 & 9.80 
& 42.9 & 38.7 & 8.90 &  72.2 & 55.7 & 19.4 & 93.6 & 82.3 & 26.8 \\
PRUNE&  
&28.5 & 15.9 & 2.60 & 81.5 & 66.7 & 7.10& 92.4 & 62.5 & 9.20 
& 43.8 & 37.5 & 10.2 &  70.4 & 55.7 & 18.8 & 94.6 & 83.3 & 26.8  \\
AdaEdit&  
& 30.5 & 18.8 & 2.90 &  79.7 & 65.7 & 7.10 & 89.4 & 58.9 & 9.70 
& 47.9 & 42.6 & 11.4 & 68.2 & 56.8 & 19.6 & 90.8 & 81.9 & 26.7  \\
SPHERE&  
& 37.6 & 25.5 & 3.30 & 78.4 & 65.8 & \underline{7.60}  & 82.4 & 54.3 & \underline{10.2} 
& 49.6 & 42.6 & 16.2&  57.7 & 46.7 & 19.6 & 81.7 & 66.7 & 26.7  \\
SimIE&  
&   28.9 & 16.7 & 3.50  &  67.2 & 61.7 & \textbf{7.90} & 79.3 & 61.7 & \textbf{10.6} 
& 44.6 & 34.2 & 16.1   &  55.5 & 41.2 & 20.2 & 78.7 & 62.3 & \underline{26.9}  \\
EMMET&  
& 54.6 &  30.5 &  3.10&\underline{86.5} & \textbf{72.2} & 5.60 & \underline{96.5} & 67.3 & 7.20 
& 64.3&  58.8 &  16.5 &  79.2 & 61.6 & 20.4 & 85.1 & 71.0 & 22.4 \\
HSE&   
& 26.7 & 16.0 & 2.40 & 82.4 & 69.7 & 6.60 & 96.4 & \underline{67.9} & 8.60 
& 38.9 & 35.1 & 6.20 & 75.2 & 56.7 & 16.6 & \textbf{95.6} & \textbf{86.0} & 26.4 \\
AlphaEdit& 
& \underline{83.7} &\underline{41.1} & \underline{3.60} & 83.5 & 43.5 &3.80 & 82.1 & 44.9 & 3.80 
& \underline{83.5} & \underline{75.6} & \underline{24.5} & 82.9 & 69.8 & \underline{23.5} & 79.2 & 66.2 & 22.6 \\

\textbf{\textsc{GEdit}}& 
& \textbf{89.8} & \textbf{51.0} & \textbf{6.10} & \textbf{90.6} & \underline{71.8} & 6.80 & \textbf{96.6} & \textbf{70.8} & 9.10 
& \textbf{88.5} & \textbf{80.2} & \textbf{26.6}& \textbf{92.9} & \textbf{81.3} & \textbf{26.1} & \underline{94.6} & \underline{85.1} & \textbf{27.2} \\
\midrule
Pre-edited &&0.30&0.40&21.6& 0.30 & 0.40 & 21.6&0.30&0.40&21.6& 0.00 & 0.00 & 37.2& 0.00 & 0.00 & 37.2& 0.00 & 0.00 & 37.2 \\
\midrule
FT& \multirow{11}{*}{\rotatebox{90}{LLaMA3-8B-Instruct}} 
& 0.00 &  0.00 &  0.00 & 0.00 & 0.00 & 0.00 & 4.10 & 1.60 & 2.50 
&0.00 &  0.00 &  0.00  & 0.00 & 0.00 & 0.00 & 4.40 & 2.20 & 6.10 \\
RLEdit&  
& 0.00 &  0.00 &  0.00 &64.2 & 48.8 &4.90 &  60.7 & 42.7 &6.50  
& 0.00 & 0.00 & 0.00 & 67.4 & 52.2 & 22.3 &  52.8 & 38.8 & 28.5 \\
RECT&   
& 0.00 &  0.00 &  0.00& 69.9 & 54.6 & 9.10  & 94.1 & 65.2 & 12.5 
& 88.5 & 75.6 & 35.6 & 85.3 & 72.6 & 35.2&  89.1 & 78.7 & \underline{36.5} \\
PRUNE&  
& 0.00 &  0.00 &  0.00 &72.6 & 55.6 & 8.80  & 92.3 & 64.3 & 12.4 
& 88.8 & 73.3 & 34.7 &86.1 & 74.6 & 34.2 & 92.1 & \underline{83.3} & 35.9\\
AdaEdit&  
& 0.00 &  0.00 &  0.00 &70.8 & 56.9 & 9.00  &  90.5 & 62.7 & 13.3
& 83.7 & 70.3 & 34.8 &81.0 & 70.1 & 35.2&  87.5 & 74.1 & 36.1\\
SPHERE&  
&\underline{41.3} & \underline{25.7} & \underline{5.30} & 67.9 & 53.8 & 9.50  & 84.6 & 59.5 & \textbf{14.0} 
& 86.5 & 71.7 & \underline{35.9} & 79.7 & 66.8 & \underline{36.0} & 85.7 & 72.7 & 36.1 \\
SimIE&  
&  24.6 & 15.5 & 3.70 &54.6 & 41.6 & \underline{10.8} & 81.3 & 55.7 & \underline{13.3} 
& 78.6&66.3 & \textbf{36.1} &81.2 & 69.3 & \textbf{36.2} &  81.8 & 66.7 & \textbf{37.5} \\
EMMET&  
& 0.00 &  0.00 &  0.00& 0.00 & 0.00 & 0.00 &  90.1 & 62.8 & 9.50 
& 0.00 &  0.00 &  0.00 & 83.7 & 71.5 & 28.2&  \textbf{94.6} & \textbf{84.5} & 31.5\\
HSE&   
& 0.00 &  0.00 &  0.00 & \underline{73.7} & \underline{57.6} & 8.40 & \textbf{94.3} & \underline{65.6} & 12.0 
& \underline{90.2}&\underline{82.1} & 35.2 & \underline{87.7} & \underline{76.6} & 35.2 & 93.0 & 83.0 & 35.5 \\
AlphaEdit& 
& 0.00 &  0.00 &  0.00 & 0.00 &  0.00 &  0.00 & 76.8 & 56.2 & 10.4 
&0.00&0.00 & 0.00 & 81.4 & 68.5 & 30.2 & 88.4 & 78.1 & 34.2 \\
\textbf{\textsc{GEdit}}& 
& \textbf{81.8} & \textbf{64.7} & \textbf{11.8} & \textbf{86.6} & \textbf{67.8} & \textbf{12.7} & \underline{92.6} & \textbf{71.8} & 13.0 
& \textbf{91.5} & \textbf{83.6} & 35.6& \textbf{92.9} & \textbf{82.1} &35.8 & \underline{93.5} & 81.1 & 36.4 \\
\bottomrule
\end{tabular}%
}
\end{table*}
\subsection{Editing Performance (RQ1)}
Table \ref{tab:results} presents editing performance over 10,000 edits conducted with varying batch sizes. The results show that all baseline methods are highly sensitive to batch size. On the CounterFact dataset, when editing with a batch size of 1 — i.e., continuously updating the LLM 10,000 times, far exceeding the 1,000–5,000 updates typically performed in prior work \citep{guotowards,fang2025hippocampal,liu2025edit} — baselines such as HSE and AlphaEdit fail entirely on the LLaMA model: efficacy, generality, and locality all collapse to zero. This phenomenon, known as \textit{model collapse} \citep{gu2024model,gupta2024unified,fangalphaedit,thedewikibigedit}, is characterized by instability that causes pre-trained knowledge to be lost, past edits to be forgotten, and new edits to fail. However, as batch size increases, the performance of these methods improves rapidly.

To enhance editing stability, regularized sequential editors like \textsc{RECT} and \textsc{SPHERE} adopt a \textit{post-hoc} regularization strategy that introduces a second stage to preserve locality. However, this two-stage design often improves locality at the cost of editing efficacy and generality. In contrast, by performing regularization \textit{in situ}, \textsc{GEdit} jointly optimizes both efficacy and locality within a single stage, achieving superior overall performance across diverse settings. Additional experimental results—including evaluations under the single-fact editing setting \citep{meng2022locating,gu2024model}, evaluation of the model's general capability retention \citep{gu2024model}, and comparisons against an extended set of baselines \citep{yu2024melo}—are presented in Appendix~\ref{app:results}.
\begin{figure*}[t]
  \centering
  \includegraphics[width=0.95\linewidth]{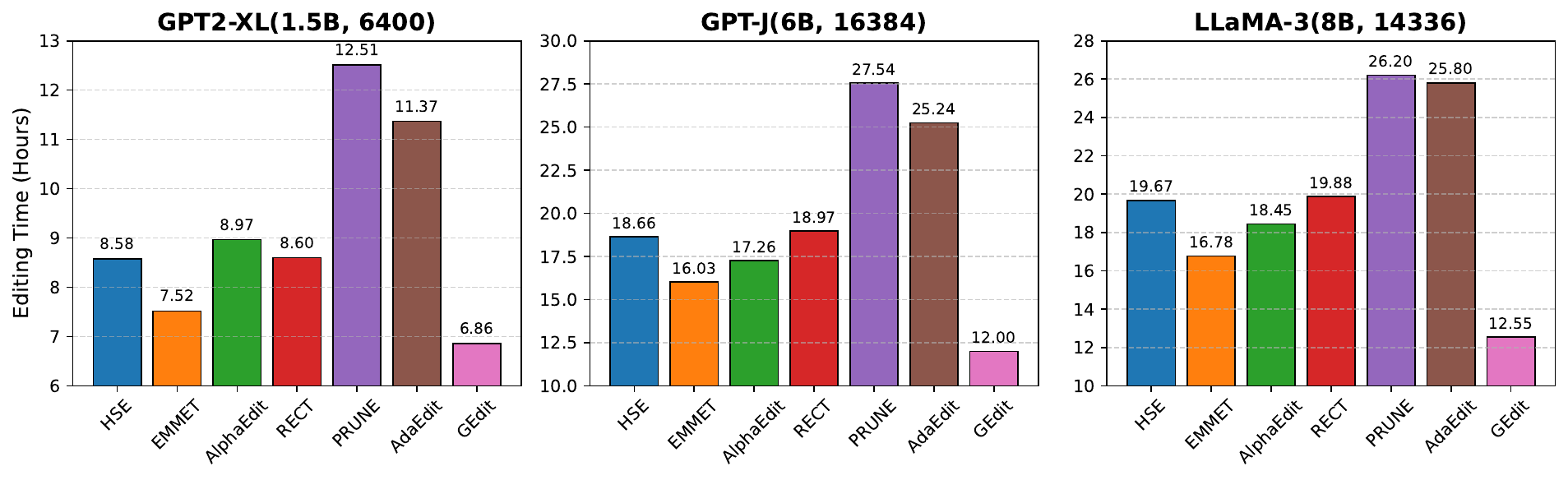}
  \caption{
    Total editing time for 10,000 sequential edits (batch size = 1) on the \textsc{CounterFact} dataset using three LLMs. From left to right: GPT2-XL, GPT-J-6B, and LLaMA3-8B, with the number of parameters and FFN hidden dimension shown in parentheses.
  }
  \label{fig:editing_time_left_only}
\end{figure*}
\begin{figure}[htbp]
    \centering
    \begin{minipage}{0.49\linewidth}
        \centering
        \includegraphics[width=1.0\linewidth]{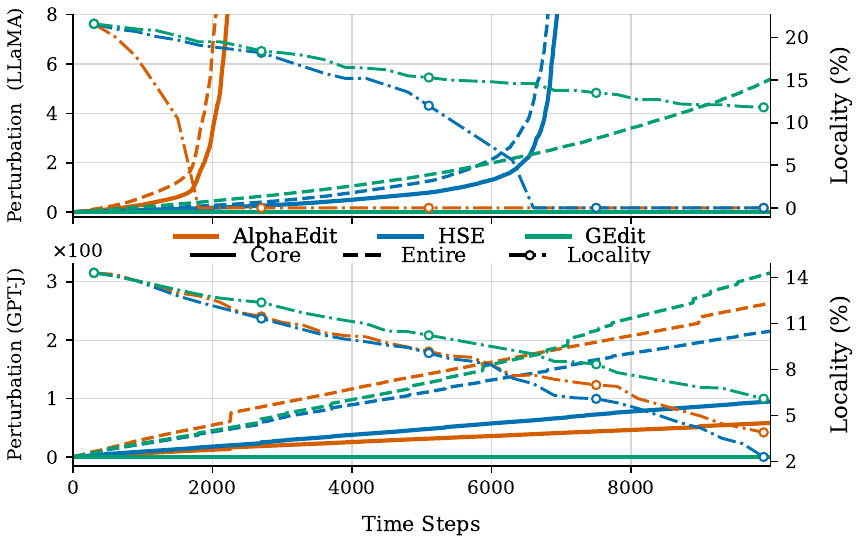}
\caption{Evolution of knowledge perturbation (entire space vs. core subspace) and edit locality during 10,000 sequential edits.}
        \label{fig:perturb}
    \end{minipage}
    \hfill
    \begin{minipage}{0.49\linewidth}
        \centering
        \includegraphics[width=1.0\linewidth]{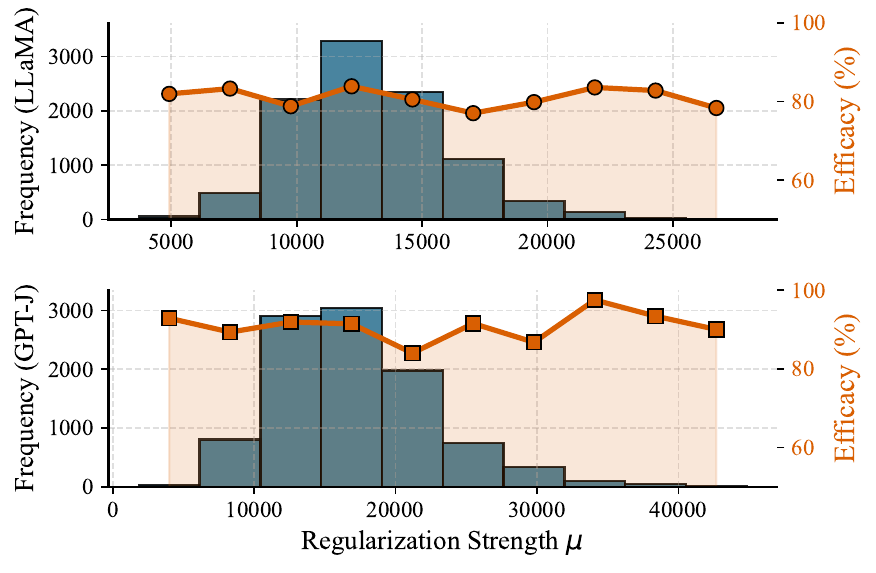}
        \caption{Distribution of the adaptive regularization strength across 10,000 sequential edits, along with the corresponding edit efficacy per bin.}
        \label{fig:mu}
    \end{minipage}
\end{figure}

\subsection{Impact of Low-Rank Regularization (RQ2)}
Low-rank regularization aims to identify a core knowledge subspace $\mathbf{U}$ and apply targeted protection, thereby confining edits to the non-core knowledge subspace and preserving locality. To demonstrate this, we measure, at each editing step $t$, the perturbation within the entire knowledge space $\|\Delta_t \mathbf{K}_0\|_F^2$ and within the core knowledge subspace $\|\Delta_t \mathbf{U}\|_F^2$, where $\Delta_t = \mathbf{W}_t - \mathbf{W}_0$ ($\mathbf{W}_t$ and $\mathbf{W}_0$ denote the weights after step $t$ and before editing, respectively). Figure~\ref{fig:perturb} compares the knowledge perturbation induced by \textsc{HSE}, AlphaEdit, and \textsc{GEdit} when editing the CounterFact dataset; results on the ZsRE dataset are provided in Appendix~\ref{subapp:low_adapt}.

Our results reveal an interesting finding: a larger $\|\Delta_t \mathbf{K}_0\|_F^2$ does not necessarily imply worse locality. Specifically, \textsc{GEdit} exhibits a larger $\|\Delta_t \mathbf{K}_0\|_F^2$ than \textsc{HSE} in the early stages when editing LLaMA, and in the later stages when editing GPT-J. Nevertheless, \textsc{GEdit} consistently achieves higher locality than \textsc{HSE} across these scenarios. We attribute this counterintuitive phenomenon to \textsc{GEdit}'s strict protection of the core knowledge subspace, where $\|\Delta_t \mathbf{U}\|_F^2 = 0$. Because this subspace encapsulates a high density of factual knowledge, rigorously preserving it is particularly beneficial for maintaining edit locality and stability.

Beyond knowledge preservation, low-rank regularization also offers a significant computational advantage. Standard locate-then-edit (LTE) methods involve a matrix inversion with $\mathcal{O}(d^3)$ cost, which is prohibitive for large $d$ and especially pronounced during multi-layer and sequential editing. By acquiring a low-rank structure, our geometry-aware regularization effectively reduces this complexity to $\mathcal{O}((d-r)^3)$, enabling highly efficient computation. As shown in Figure~\ref{fig:editing_time_left_only}, \textsc{GEdit} significantly accelerates the 10,000 sequential edits on \textsc{CounterFact}, completing the process for GPT-J and LLaMA in only \textbf{12 hours}—substantially faster than baselines such as \textsc{HSE} and \textsc{AlphaEdit}.\footnote{All experiments were conducted on an NVIDIA A6000 (48GB). All evaluated methods use LLMs loaded in \texttt{bfloat16} precision for faster editing.}

\begin{wraptable}[9]{r}{0.43\textwidth}
\caption{Effect of low-rank regularization on different editing methods.}
\label{tab:ada_lr}
\setlength{\tabcolsep}{2pt}
\scriptsize
\begin{tabular}{l c c c c}
\toprule
\multirow{2}{*}{Method} & 
\multicolumn{2}{c}{\textbf{GPT-J}} & 
\multicolumn{2}{c}{\textbf{LLaMA}} \\
\cmidrule(lr){2-3} \cmidrule(lr){4-5}
& \textbf{Eff.} & \textbf{Loc.} & \textbf{Eff.} & \textbf{Loc.} \\
\midrule
EMMET       & 54.6->76.4 & 3.1->4.3& (2k)->(4k) & (2k)->(4k) \\
HSE       & 26.7->71.7 & 2.4->4.7 & (7k)->(10k) & (7k)->(10k) \\
AlphaEdit & 83.7->85.4 & 3.6->4.2 & 0.0->77.8 & 0.0->10.1 \\
\bottomrule
\end{tabular}
\end{wraptable}
The low-rank regularization is general and can be applied to various editing methods by simply replacing the knowledge preservation term with the expected output perturbation term under the LR+D factor model. Table~\ref{tab:ada_lr} demonstrates its effectiveness when integrated into EMMET, HSE, and AlphaEdit, with arrows indicating performance changes over 10,000 sequential edits on the Counterfact dataset. For GPT-J, all methods show improvements in both efficacy and locality. For LLaMA, since EMMET and HSE achieve zero Eff. and Loc. regardless of whether low-rank regularization is applied, we report the collapse step at which performance first drops to 0 (evaluated every 1,000 steps). Details on adapting these methods to low-rank regularization and additional experimental results are provided in Appendix~\ref{subapp:low_adapt}.

\subsection{Impact of Adaptive Regularization (RQ3)}

Our adaptive regularization framework, grounded in Theorem~\ref{th:ee}, provides a unified way to adjust regularization strength across edits, models, datasets, and methods via editability, as it is agnostic to these factors. In contrast, as demonstrated in Appendix~\ref{subapp:ada_adapt}, editing performance with fixed regularization is highly sensitive to such factors and—even when carefully tuned—consistently underperforms its adaptive counterpart in balancing edit efficacy and knowledge preservation.

\begin{wraptable}[10]{r}{0.43\textwidth}
\caption{Effect of adaptive regularization on different editing methods.}
\label{tab:ada_r}
\setlength{\tabcolsep}{2pt}
\scriptsize
\begin{tabular}{l c c c c}
\toprule
\multirow{2}{*}{Method} & 
\multicolumn{2}{c}{\textbf{GPT-J}} & 
\multicolumn{2}{c}{\textbf{LLaMA}} \\
\cmidrule(lr){2-3} \cmidrule(lr){4-5}
& \textbf{Eff.} & \textbf{Loc.} & \textbf{Eff.} & \textbf{Loc.} \\
\midrule
EMMET       & 54.6->77.5 & 3.1->4.5 & 0.0->70.3 & 0.0->6.8 \\
HSE       & 26.7->78.1 & 2.4->4.2 & 0.0->72.3 & 0.0->7.4 \\
AlphaEdit & 83.7->84.1 & 3.6->5.7 & (2k)->(5k) & (2k)->(5k) \\
\bottomrule
\end{tabular}
\end{wraptable}
To demonstrate the effectiveness of the adaptive regularization across edits, we record its regularization strength at each step during 10,000 sequential edits on the CounterFact dataset and plot its frequency distribution in Figure~\ref{fig:mu}; results on the ZsRE dataset are provided in Appendix~\ref{subapp:ada_adapt}. The results show that high editing efficacy can be maintained across a wide range of regularization strengths. To demonstrate the effectiveness of the adaptive regularization across methods, we apply it to EMMET, HSE, and AlphaEdit using a \textit{uniform} desired editability $c = 1.25$, despite their default regularization strengths varying drastically (e.g., 10 for AlphaEdit vs. 15,000 for HSE). Table~\ref{tab:ada_r} reports the performance of EMMET, HSE, and AlphaEdit equipped with our adaptive regularization, evaluated over 10,000 sequential edits on the \textsc{CounterFact} dataset. The results show that our approach consistently and substantially outperforms the fixed-regularization counterparts.


\section{Conclusion}
\label{sec:conclusion}

We presented \textsc{GEdit}, a geometry-aware knowledge editing framework with two complementary components: low-rank regularization and adaptive regularization. The former identifies a low-rank core knowledge subspace for targeted protection, while the latter quantifies the editability of individual edit requests and adjusts regularization strength accordingly. Together, they achieve a better balance between editing efficacy and knowledge preservation. Extensive experiments demonstrate the effectiveness and robustness of \textsc{GEdit} across diverse models, editing methods, datasets, and batch sizes, including large-scale sequential editing with up to 10,000  updates.

Our work also has limitations that warrant further investigation. First, our theoretical analysis characterizes the relationship between editability and editing success for a single edit, but does not directly extend to jointly optimizing multiple edits, where conflicts among edit requests may affect the realized output perturbations \citep{dai2025eamet}. Although averaging individual editability scores may provide a summary of batch-level editability, it does not guarantee the same theoretical relationship for batch editing. Extending the analysis to characterize such conflicts and establish theoretical guarantees for batch editing is an important direction for future work. Second, our current framework uses a globally selected rank for the core knowledge subspace, which does not account for the geometric characteristics of individual edit requests. Developing an edit-adaptive rank selection mechanism may further improve the trade-off between editing efficacy and knowledge preservation.

\section*{Acknowledgments}
This work was supported by the National Key Research and Development Program of China under Grant 2024YFC3307900; the National Natural Science Foundation of China under Grants 62376103, 62302184, 62436003, 62472332, and 62672205; the Major Science and Technology Project of Hubei Province under Grants 2025BAB011, 2024BAA008, and 2024BAB039; the Hubei Science and Technology Talent Service Project under Grant 2024DJC078; the Science and Technology Innovation Talent Program under grant 2023CB005; and Wuhan Dameng Database Co., Ltd.

\bibliographystyle{plainnat}
\bibliography{ref}

\appendix
\section{Modeling the FFN Hidden Representations via LR+D Factor Model}
\label{app:lrd}

\subsection{Covariance Structure and the Expected Regularization Term}
\label{subapp:property}

We derive the covariance structure of FFN hidden representations and the corresponding expected output perturbation under the Low-Rank Plus Diagonal (LR+D) factor model introduced in Section~\ref{sec:method}.

Specifically, we assume that a hidden vector $\mathbf{k}\in\mathbb{R}^d$ follows the generative model
\begin{equation}
    \mathbf{k} = \mathbf{b} + \mathbf{U}\mathbf{z} + \boldsymbol{\varepsilon},
    \label{eq:factor_model_app}
\end{equation}
where $\mathbf{b}=\mathbb{E}[\mathbf{k}]\in\mathbb{R}^d$ is the mean vector, $\mathbf{z}\sim\mathcal{N}(\mathbf{0},\mathbf{I}_r)$ is a latent semantic code, $\mathbf{U}\in\mathbb{R}^{d\times r}$ spans the dominant semantic subspace, and $\boldsymbol{\varepsilon}\sim\mathcal{N}(\mathbf{0},\mathbf{D})$ captures dimension-wise residual variation. Here, $\mathbf{D}$ is a diagonal  matrix, and $\mathbf{z}$ and $\boldsymbol{\varepsilon}$ are independent.

Since both $\mathbf{z}$ and $\boldsymbol{\varepsilon}$ have zero means, we have
\begin{equation}
    \mathbb{E}[\mathbf{k}]=\mathbf{b}.
\end{equation}
The covariance matrix of $\mathbf{k}$ is therefore
\begin{align}
    \mathrm{Cov}(\mathbf{k})
    &= \mathbb{E}\left[
        (\mathbf{k}-\mathbf{b})(\mathbf{k}-\mathbf{b})^\top
    \right] \nonumber\\
    &= \mathbb{E}\left[
        (\mathbf{U}\mathbf{z}+\boldsymbol{\varepsilon})
        (\mathbf{U}\mathbf{z}+\boldsymbol{\varepsilon})^\top
    \right] \nonumber\\
    &= \mathbf{U}\mathbb{E}[\mathbf{z}\mathbf{z}^\top]\mathbf{U}^\top
       + \mathbb{E}[
           \boldsymbol{\varepsilon}\boldsymbol{\varepsilon}^\top
       ] \nonumber\\
    &\quad
       + \mathbf{U}\mathbb{E}[
           \mathbf{z}\boldsymbol{\varepsilon}^\top
       ]
       + \mathbb{E}[
           \boldsymbol{\varepsilon}\mathbf{z}^\top
       ]\mathbf{U}^\top \nonumber\\
    &= \mathbf{U}\mathbf{U}^\top+\mathbf{D}.
    \label{eq:cov_lrd}
\end{align}

Unlike the covariance, the second moment of $\mathbf{k}$ also accounts for its nonzero mean. Specifically,
\begin{align}
    \mathbb{E}[\mathbf{k}\mathbf{k}^\top]
    &= \mathrm{Cov}(\mathbf{k})
       + \mathbb{E}[\mathbf{k}]
         \mathbb{E}[\mathbf{k}]^\top \nonumber\\
    &= \mathbf{U}\mathbf{U}^\top
       + \mathbf{D}+\mathbf{b}\mathbf{b}^\top.
    \label{eq:second_moment_lrd}
\end{align}
For convenience, define
\begin{equation}
    \bar{\mathbf{D}}
    := \mathbf{b}\mathbf{b}^\top+\mathbf{D}.
    \label{eq:barD_app}
\end{equation}
Then
\begin{equation}
    \mathbb{E}[\mathbf{k}\mathbf{k}^\top]
    = \mathbf{U}\mathbf{U}^\top+\bar{\mathbf{D}}.
\end{equation}

We next derive the expected output perturbation induced by an edit matrix $\Delta$. Since the original output is $\mathbf{v}=\mathbf{W}\mathbf{k}$, the output perturbation is
\begin{equation}
    (\mathbf{W}+\Delta)\mathbf{k}-\mathbf{v}
    =\Delta\mathbf{k}.
\end{equation}
Consequently,
\begin{align}
    \mathbb{E}_{\mathbf{k}}
    \left[\|\Delta\mathbf{k}\|_2^2\right]
    &= \mathbb{E}_{\mathbf{k}}
       \left[\mathbf{k}^\top\Delta^\top\Delta\mathbf{k}\right]
       \nonumber\\
    &= \mathrm{Tr}\left(
        \Delta^\top\Delta
        \mathbb{E}[\mathbf{k}\mathbf{k}^\top]
       \right) \nonumber\\
    &= \mathrm{Tr}\left(
        \Delta^\top\Delta\mathbf{U}\mathbf{U}^\top
       \right)
       + \mathrm{Tr}\left(
        \Delta^\top\Delta\bar{\mathbf{D}}
       \right).
    \label{eq:exp_perturbation_app}
\end{align}
The second equality follows from the identity
$\mathbf{x}^\top\mathbf{A}\mathbf{x}
=\mathrm{Tr}(\mathbf{A}\mathbf{x}\mathbf{x}^\top)$
and the linearity of expectation and trace.

Equation~\eqref{eq:exp_perturbation_app} provides the theoretical foundation for the geometry-aware regularization in Section~\ref{sec:method}. The first term penalizes changes induced by the dominant semantic subspace, whereas the second accounts for the remaining second-order contributions, including the residual variation and the nonzero mean of the hidden representations.
\subsection{Closed-Form Solution of the LR+D Regularized Editing Objective}
\label{subapp:closed_form}

We derive the closed-form solution to the optimization problem in Equation~\ref{eq:lr_obj}:
\begin{equation}
    \min_{\Delta}\quad
    \left\|(\mathbf{W}+\Delta)\mathbf{K}_1-\mathbf{V}_1\right\|_F^2
    +\beta\,\mathrm{Tr}\left(\Delta^\top\Delta\mathbf{U}\mathbf{U}^\top\right)
    +\mu\,\mathrm{Tr}\left(\Delta^\top\Delta\bar{\mathbf{D}}\right).
    \label{eq:app_objective}
\end{equation}

Define the residual matrix
\begin{equation}
    \mathbf{R}_1:=\mathbf{V}_1-\mathbf{W}\mathbf{K}_1.
    \label{eq:residual_app}
\end{equation}
The objective can then be rewritten as
\begin{equation}
    \mathcal{L}(\Delta)
    =\|\Delta\mathbf{K}_1-\mathbf{R}_1\|_F^2
    +\beta\,\mathrm{Tr}\left(\Delta^\top\Delta\mathbf{U}\mathbf{U}^\top\right)
    +\mu\,\mathrm{Tr}\left(\Delta^\top\Delta\bar{\mathbf{D}}\right).
    \label{eq:app_objective_residual}
\end{equation}

Expanding the Frobenius norm yields
\begin{equation}
    \|\Delta\mathbf{K}_1-\mathbf{R}_1\|_F^2
    =
    \mathrm{Tr}\left(\Delta\mathbf{K}_1\mathbf{K}_1^\top\Delta^\top\right)
    -2\,\mathrm{Tr}\left(\mathbf{R}_1\mathbf{K}_1^\top\Delta^\top\right)
    +\mathrm{Tr}\left(\mathbf{R}_1\mathbf{R}_1^\top\right).
    \label{eq:app_frobenius_expansion}
\end{equation}

Using the cyclic property of the trace and omitting terms independent of $\Delta$, we obtain
\begin{equation}
    \mathcal{L}(\Delta)
    =
    \mathrm{Tr}\left(\Delta^\top\Delta\mathbf{K}_1\mathbf{K}_1^\top\right)
    -2\,\mathrm{Tr}\left(\Delta^\top\mathbf{R}_1\mathbf{K}_1^\top\right)
    +\beta\,\mathrm{Tr}\left(\Delta^\top\Delta\mathbf{U}\mathbf{U}^\top\right)
    +\mu\,\mathrm{Tr}\left(\Delta^\top\Delta\bar{\mathbf{D}}\right)
    +\mathrm{const}.
    \label{eq:app_expanded_objective}
\end{equation}

Taking the derivative with respect to $\Delta$ and using the standard matrix calculus identities
\begin{equation}
    \nabla_{\Delta}\,\mathrm{Tr}(\Delta^\top\Delta\mathbf{A})=2\Delta\mathbf{A},
    \qquad
    \nabla_{\Delta}\,\mathrm{Tr}(\Delta^\top\mathbf{B})=\mathbf{B},
\end{equation}
for symmetric $\mathbf{A}$, we obtain
\begin{equation}
    \nabla_{\Delta}\mathcal{L}
    =
    2\Delta\mathbf{K}_1\mathbf{K}_1^\top
    -2\mathbf{R}_1\mathbf{K}_1^\top
    +2\beta\Delta\mathbf{U}\mathbf{U}^\top
    +2\mu\Delta\bar{\mathbf{D}}.
    \label{eq:app_gradient}
\end{equation}
Setting the gradient to zero gives the first-order optimality condition
\begin{equation}
    \Delta\left(
        \mathbf{K}_1\mathbf{K}_1^\top
        +\beta\mathbf{U}\mathbf{U}^\top
        +\mu\bar{\mathbf{D}}
    \right)
    =\mathbf{R}_1\mathbf{K}_1^\top.
    \label{eq:app_optimality}
\end{equation}
Solving for $\Delta$ yields the closed-form solution in Equation~\ref{eq:update_combined}:
\begin{equation}
    \Delta
    =\mathbf{R}_1\mathbf{K}_1^\top
    \left(
        \mathbf{K}_1\mathbf{K}_1^\top
        +\beta\mathbf{U}\mathbf{U}^\top
        +\mu\bar{\mathbf{D}}
    \right)^{-1}.
    \label{eq:app_closed_form}
\end{equation}

\subsection{Proof of the Null-Space Limit}
\label{app:alphaedit_conn}

We aim to prove that
\[
\lim_{\beta \to \infty} \left( \mathbf{K}_1 \mathbf{K}_1^\top + \beta \mathbf{U} \mathbf{U}^\top + \mu \bar{\mathbf{D}} \right)^{-1} = \mathbf{M} \left( \mathbf{M}^\top (\mathbf{K}_1 \mathbf{K}_1^\top + \mu \bar{\mathbf{D}}) \mathbf{M} \right)^{-1} \mathbf{M}^\top,
\]
where the columns of \(\mathbf{M}\) form an orthonormal basis for the orthogonal complement of \(\operatorname{range}(\mathbf{U})\).

Let \(\mathbf{P}_{\parallel} = \mathbf{U}(\mathbf{U}^\top\mathbf{U})^{-1}\mathbf{U}^\top\) be the orthogonal projector onto \(\mathcal{S} = \operatorname{range}(\mathbf{U})\), and let \(\mathbf{P}_{\perp} = \mathbf{I} - \mathbf{P}_{\parallel} = \mathbf{M}\mathbf{M}^\top\) be the projector onto \(\mathcal{S}^\perp\). Since \(\{\mathbf{U}, \mathbf{M}\}\) forms a complete orthogonal basis for \(\mathbb{R}^d\), we can express any matrix in this basis. In particular, we can write the matrix \(\mathbf{A}_\beta := \mathbf{K}_1 \mathbf{K}_1^\top + \beta \mathbf{U} \mathbf{U}^\top + \mu \bar{\mathbf{D}}\) in block form with respect to the decomposition \(\mathbb{R}^d = \mathcal{S} \oplus \mathcal{S}^\perp\):
\begin{equation}
    \mathbf{A}_\beta =
    \begin{bmatrix}
        \mathbf{P}_{\parallel} \mathbf{A}_\beta \mathbf{P}_{\parallel} & \mathbf{P}_{\parallel} \mathbf{A}_\beta \mathbf{P}_{\perp} \\
        \mathbf{P}_{\perp} \mathbf{A}_\beta \mathbf{P}_{\parallel} & \mathbf{P}_{\perp} \mathbf{A}_\beta \mathbf{P}_{\perp}
    \end{bmatrix}
    =
    \begin{bmatrix}
        \beta \mathbf{U}^\top\mathbf{U} + \mathbf{C}_{11} & \mathbf{C}_{12} \\
        \mathbf{C}_{21} & \mathbf{C}_{22}
    \end{bmatrix},
\end{equation}
where the blocks \(\mathbf{C}_{ij}\) (which depend on \(\mathbf{K}_1, \bar{\mathbf{D}}, \mu\) but not on \(\beta\)) are defined as:
\begin{align}
    \mathbf{C}_{11} &= \mathbf{U}^\top (\mathbf{K}_1 \mathbf{K}_1^\top + \mu \bar{\mathbf{D}}) \mathbf{U}, \\
    \mathbf{C}_{12} &= \mathbf{U}^\top (\mathbf{K}_1 \mathbf{K}_1^\top + \mu \bar{\mathbf{D}}) \mathbf{M}, \\
    \mathbf{C}_{21} &= \mathbf{M}^\top (\mathbf{K}_1 \mathbf{K}_1^\top + \mu \bar{\mathbf{D}}) \mathbf{U}, \\
    \mathbf{C}_{22} &= \mathbf{M}^\top (\mathbf{K}_1 \mathbf{K}_1^\top + \mu \bar{\mathbf{D}}) \mathbf{M}.
\end{align}
Note that \(\mathbf{C}_{22}\) is symmetric positive definite because \(\mathbf{K}_1 \mathbf{K}_1^\top + \mu \bar{\mathbf{D}} \succ 0\).

Now, consider the inverse \(\mathbf{A}_\beta^{-1}\). Using the standard formula for the inverse of a \(2\times2\) block matrix, its bottom-right block (which corresponds to the action of the inverse on \(\mathcal{S}^\perp\)) is given by the Schur complement:
\begin{equation}
    (\mathbf{A}_\beta^{-1})_{22} = \left( \mathbf{C}_{22} - \mathbf{C}_{21} (\beta \mathbf{U}^\top\mathbf{U} + \mathbf{C}_{11})^{-1} \mathbf{C}_{12} \right)^{-1}.
\end{equation}
As \(\beta \to \infty\), the term \((\beta \mathbf{U}^\top\mathbf{U} + \mathbf{C}_{11})^{-1} \to \mathbf{0}\), because \(\mathbf{U}^\top\mathbf{U}\) is positive definite. Therefore,
\begin{equation}
    \lim_{\beta \to \infty} (\mathbf{A}_\beta^{-1})_{22} = \mathbf{C}_{22}^{-1} = \left( \mathbf{M}^\top (\mathbf{K}_1 \mathbf{K}_1^\top + \mu \bar{\mathbf{D}}) \mathbf{M} \right)^{-1}.
\end{equation}
Furthermore, all other blocks of \(\mathbf{A}_\beta^{-1}\) (i.e., \((\mathbf{A}_\beta^{-1})_{11}, (\mathbf{A}_\beta^{-1})_{12}, (\mathbf{A}_\beta^{-1})_{21}\)) contain factors of \((\beta \mathbf{U}^\top\mathbf{U} + \mathbf{C}_{11})^{-1}\) and thus vanish in the limit \(\beta \to \infty\).

Consequently, the full limiting matrix is zero everywhere except on the \(\mathcal{S}^\perp\) subspace, where it acts as \(\mathbf{C}_{22}^{-1}\). This is precisely expressed in the original coordinate system as:
\begin{equation}
    \lim_{\beta \to \infty} \mathbf{A}_\beta^{-1} = \mathbf{M} \left( \mathbf{M}^\top (\mathbf{K}_1 \mathbf{K}_1^\top + \mu \bar{\mathbf{D}}) \mathbf{M} \right)^{-1} \mathbf{M}^\top.
\end{equation}
Multiplying this limit from the left by the \(\beta\)-independent term \(\mathbf{R}\mathbf{K}_1^\top\) yields the desired result for \(\Delta_{\mathbf{G}}\).

\section{Geometry-Aware Adaptive Regularization via Editability}

\subsection{Derivation of Editability as Mahalanobis Distance}
\label{subapp:mahal_derivation}

We show that the editability in Equation~\eqref{eq:editability} reduces to a squared Mahalanobis distance when each principal direction is weighted by the inverse of its eigenvalue, i.e., $\sigma_j = 1/\lambda_j$. Let $\mathbf{C}_0=\mathbf{Q}\boldsymbol{\Lambda}\mathbf{Q}^\top$ be the eigendecomposition of the empirical second-moment matrix of the FFN hidden representations $\mathbf{k}$, where $\mathbf{Q}$ is orthogonal, $\boldsymbol{\Lambda}=\mathrm{diag}(\lambda_1,\dots,\lambda_d)$ with $\lambda_j>0$, and $\mathbf{q}_j$ denotes the $j$-th column of $\mathbf{Q}$. Substituting $\sigma_j=1/\lambda_j$ into Equation~\eqref{eq:editability} yields
\begin{align}
    \mathcal{E}(\mathbf{k}_e) &= \frac{1}{\mu} \sum_{j=1}^d \frac{1}{\lambda_j} (\mathbf{k}_e^\top \mathbf{q}_j)^2 \nonumber \\
    &= \frac{1}{\mu} \sum_{j=1}^d \frac{1}{\lambda_j} (\mathbf{q}_j^\top \mathbf{k}_e) (\mathbf{k}_e^\top \mathbf{q}_j) \nonumber \\
    &= \frac{1}{\mu} \mathbf{k}_e^\top \left( \sum_{j=1}^d \frac{1}{\lambda_j} \mathbf{q}_j \mathbf{q}_j^\top \right) \mathbf{k}_e.
\end{align}
The term in parentheses is precisely the eigendecomposition of the inverse second-moment matrix $\mathbf{C}_0^{-1}$, since
\begin{equation}
    \mathbf{C}_0^{-1}
    = (\mathbf{Q}\boldsymbol{\Lambda}\mathbf{Q}^\top)^{-1}
    = \mathbf{Q}\boldsymbol{\Lambda}^{-1}\mathbf{Q}^\top
    = \sum_{j=1}^d \frac{1}{\lambda_j} \mathbf{q}_j \mathbf{q}_j^\top.
\end{equation}
Therefore, the editability simplifies to
\begin{equation}
    \mathcal{E}(\mathbf{k}_e) = \frac{1}{\mu} \mathbf{k}_e^\top \mathbf{C}_0^{-1} \mathbf{k}_e,
\end{equation}
which is the desired result stated in Equation~\eqref{eq:editability_c}.
\subsection{Restriction of Editability to the Non-Core Subspace}
\label{subapp:editability_derivation}

We detail how the editability in Equation~\eqref{eq:editability_c} is defined on the non-core knowledge subspace $\mathbf{M}$ under the LR+D model of Section~\ref{sec:method}. Recall that $\mathbf{M}\in\mathbb{R}^{d\times(d-r)}$ has orthonormal columns spanning the orthogonal complement of the core subspace $\mathbf{U}$, i.e., $\mathbf{M}^\top\mathbf{U}=\mathbf{0}$ and $\mathbf{M}^\top\mathbf{M}=\mathbf{I}_{d-r}$.

Under the LR+D model, the second-moment matrix of $\mathbf{k}$ is
\begin{equation}
    \mathbf{C}_0 = \mathbb{E}[\mathbf{k}\mathbf{k}^\top]
    = \mathbf{U}\mathbf{U}^\top + \bar{\mathbf{D}},
    \qquad
    \bar{\mathbf{D}} = \mathbf{b}\mathbf{b}^\top + \mathbf{D},
    \label{eq:C0_lrd_app}
\end{equation}
Since $\mathbf{M}^\top\mathbf{U}=\mathbf{0}$, multiplying by $\mathbf{M}^\top$ and $\mathbf{M}$ eliminates the low-rank component:
\begin{equation}
    \mathbf{M}^\top\mathbf{C}_0\mathbf{M}
    = \mathbf{M}^\top\bar{\mathbf{D}}\mathbf{M}.
    \label{eq:MC0M_app}
\end{equation}

For an edit vector confined to the non-core subspace, i.e., $\mathbf{k}_e=\mathbf{M}\mathbf{M}^\top\mathbf{k}_e$, we define the restricted editability using the inverse of the projected second-moment matrix:
\begin{equation}
    \mathcal{E}(\mathbf{k}_e)
    = \frac{1}{\mu}\mathbf{k}_e^\top\mathbf{M}
      \big(\mathbf{M}^\top\mathbf{C}_0\mathbf{M}\big)^{-1}
      \mathbf{M}^\top\mathbf{k}_e,
    \label{eq:editability_restricted_app}
\end{equation}
where $\mathbf{M}^\top\mathbf{C}_0\mathbf{M}$ is invertible. Substituting Equation~\eqref{eq:MC0M_app} into Equation~\eqref{eq:editability_restricted_app} yields
\begin{equation}
    \mathcal{E}(\mathbf{k}_e)
    = \frac{1}{\mu}\mathbf{k}_e^\top\mathbf{M}
      \big(\mathbf{M}^\top\bar{\mathbf{D}}\mathbf{M}\big)^{-1}
      \mathbf{M}^\top\mathbf{k}_e,
    \label{eq:editability_d_app}
\end{equation}
which is the restricted editability used in Equation~\eqref{eq:editability_d}.

\subsection{Theoretical Relationship between Editability and Edit Success}
\label{subapp:theorem_editability}

We provide a formal derivation of Theorem \ref{th:ee}.
According to Equation~\eqref{eq:gedit_null_limit}, the output perturbation on the edit input $\mathbf{k}_e$ is:
\begin{equation}
    \Delta_{\mathbf{G}} \mathbf{k}_e = \mathbf{r}_e \underbrace{\mathbf{k}_e^\top \mathbf{M} \big( \mathbf{M}^{\top}(\mathbf{k}_e \mathbf{k}_e^\top + \mu \bar{\mathbf{D}})\mathbf{M} \big)^{-1}\mathbf{M}^{\top} \mathbf{k}_e}_{S}.
\end{equation}
Our task is to simplify the scalar $S$.

Since the edit is constrained to the editable subspace, $\mathbf{k}_e \in \mathrm{span}(\mathbf{M})$, which implies $\mathbf{k}_e = \mathbf{M}\mathbf{M}^\top\mathbf{k}_e$. Let us define $\tilde{\mathbf{k}}_e = \mathbf{M}^\top \mathbf{k}_e$, which is the coordinate vector of $\mathbf{k}_e$ in the basis $\mathbf{M}$. Using this, the scalar $S$ can be rewritten as:
\begin{equation}
    S = \tilde{\mathbf{k}}_e^\top \big( \tilde{\mathbf{k}}_e \tilde{\mathbf{k}}_e^\top + \mu \mathbf{M}^\top\bar{\mathbf{D}}\mathbf{M} \big)^{-1} \tilde{\mathbf{k}}_e.
\end{equation}

Now, we apply the Sherman-Morrison-Woodbury (SMW) formula to the inverse. For an invertible matrix $\mathbf{A}$ and vector $\mathbf{u}$, the SMW identity for a rank-1 update states:
\[
(\mathbf{A} + \mathbf{u}\mathbf{u}^\top)^{-1} = \mathbf{A}^{-1} - \frac{\mathbf{A}^{-1}\mathbf{u}\mathbf{u}^\top\mathbf{A}^{-1}}{1 + \mathbf{u}^\top\mathbf{A}^{-1}\mathbf{u}}.
\]

Identifying $\mathbf{A} = \mu \mathbf{M}^\top\bar{\mathbf{D}}\mathbf{M}$ and $\mathbf{u} = \tilde{\mathbf{k}}_e$, we compute:
\begin{align}
    S &= \tilde{\mathbf{k}}_e^\top \left[ (\mu \mathbf{M}^\top\bar{\mathbf{D}}\mathbf{M})^{-1} - \frac{(\mu \mathbf{M}^\top\bar{\mathbf{D}}\mathbf{M})^{-1} \tilde{\mathbf{k}}_e \tilde{\mathbf{k}}_e^\top (\mu \mathbf{M}^\top\bar{\mathbf{D}}\mathbf{M})^{-1}}{1 + \tilde{\mathbf{k}}_e^\top (\mu \mathbf{M}^\top\bar{\mathbf{D}}\mathbf{M})^{-1} \tilde{\mathbf{k}}_e} \right] \tilde{\mathbf{k}}_e \nonumber \\
    &= \frac{1}{\mu} \tilde{\mathbf{k}}_e^\top (\mathbf{M}^\top\bar{\mathbf{D}}\mathbf{M})^{-1} \tilde{\mathbf{k}}_e - \frac{1}{\mu^2} \frac{(\tilde{\mathbf{k}}_e^\top (\mathbf{M}^\top\bar{\mathbf{D}}\mathbf{M})^{-1} \tilde{\mathbf{k}}_e)^2}{1 + \frac{1}{\mu} \tilde{\mathbf{k}}_e^\top (\mathbf{M}^\top\bar{\mathbf{D}}\mathbf{M})^{-1} \tilde{\mathbf{k}}_e} \nonumber \\
    &= \frac{\frac{1}{\mu} \tilde{\mathbf{k}}_e^\top (\mathbf{M}^\top\bar{\mathbf{D}}\mathbf{M})^{-1} \tilde{\mathbf{k}}_e}{1 + \frac{1}{\mu} \tilde{\mathbf{k}}_e^\top (\mathbf{M}^\top\bar{\mathbf{D}}\mathbf{M})^{-1} \tilde{\mathbf{k}}_e}\nonumber \\
    &=\frac{\mathcal{E}(\mathbf{k}_e)}{1 + \mathcal{E}(\mathbf{k}_e)}.
\end{align}
Therefore, the output perturbation is:
\begin{equation}
    \Delta_{\mathbf{G}} \mathbf{k}_e = \frac{\mathcal{E}(\mathbf{k}_e)}{1 + \mathcal{E}(\mathbf{k}_e)} \, \mathbf{r}_e,
\end{equation}
which completes the proof of the theorem.

\section{Experimental Details}
\label{app:exp}
\subsection{Baseline Model Editing Methods}
We summarize the core ideas of the baseline methods used in this work:

\begin{itemize}
\item \textbf{Fine-tuning based methods}: FT~\citep{hu2022lora} applies LoRA to fine-tune the weights of the second FFN linear layer. The fine-tuned layer coincides with those modified by locate-then-edit methods.

\item \textbf{Hypernetwork based methods}: These are meta-learning based editors. RLEdit~\citep{li2025reinforced} trains hypernetworks to predict localized weight updates, improving upon MEND~\citep{mitchellfast} and MALMEN~\citep{tan2024massive} in sequential editing settings.

\item \textbf{Locate-then-edit methods}:  
\textbf{HSE}~\citep{mengmass} improves MEMIT~\citep{mengmass} with history-aware residual computation and history-aware calculation of weight perturbations.  
\textbf{EMMET}~\citep{gupta2024unified} extends ROME~\citep{meng2022locating} (which only supports single-fact editing) to multi-fact scenarios.  
\textbf{AlphaEdit}~\citep{fangalphaedit} restricts weight updates to the null space of pretrained knowledge.

\item \textbf{Regularized sequential editors}: For a given weight perturbation $\Delta$:  
\textbf{RECT}~\citep{gu2024model} drops small-magnitude elements.  
\textbf{PRUNE}~\citep{maperturbation} constrains the maximum singular value and reconstructs the perturbation.  
\textbf{AdaEdit}~\citep{li2025adaedit} drops small singular directions and reconstructs the perturbation.  
\textbf{SIMIE}~\citep{guotowards} protects the weight perturbation using a projector that collects the covariance of historical edits.  
\textbf{SPHERE}~\citep{liu2026energyregularized} protects the weight perturbation using a projector that collects the principal space of the initial pretrained weights.
\end{itemize}

\subsection{Dataset Examples and Evaluation Metrics}
\label{app:data_metric}

\paragraph{Dataset Examples}
In CounterFact~\citep{mitchell2022memory}, each edit request is a quadruple $(s, r, o_{\text{ori}}, o_{\text{tar}})$ representing a factual update. An example is shown below:

\begin{itemize}
    \item \textbf{Edit request:} (Danielle Darrieux, \textit{mother tongue of}, French, English)
    \item \textbf{Edit prompt:} ``The mother tongue of Danielle Darrieux is''
    \item \textbf{Paraphrase prompts:}
    \begin{itemize}
        \item ``Danielle Darrieux, a native''
        \item ``Danielle Darrieux spoke the language''
    \end{itemize}
    \item \textbf{Neighborhood prompts (unrelated subjects, same relation and object):}
    \begin{itemize}
        \item ``The mother tongue of Léon Blum is''
        \item ``The native language of Montesquieu is''
        \item ``François Bayrou, a native''
        \item ``The native language of Raymond Barre is''
        \item ``Michel Rocard is a native speaker of''
        \item $\dots$ (10 total)
    \end{itemize}
\end{itemize}

The model should output ``English'' for the edit and paraphrase prompts after editing, while continuing to answer neighborhood prompts correctly (e.g., ``French'' for Léon Blum).

In ZsRE~\citep{levy2017zero}, edit requests are constructed from question--answer pairs. An example is shown below:

\begin{itemize}
    \item \textbf{Edit request:} (Watts Humphrey, \textit{attend university}, None, Illinois Institute of Technology)
    \item \textbf{Edit prompt:} ``What university did Watts Humphrey attend?''
    \item \textbf{Paraphrase prompts:}
    \begin{itemize}
        \item ``What university did Watts Humphrey take part in?''
    \end{itemize}
    \item \textbf{Neighborhood queries and ground-truth answers:}
    \begin{itemize}
        \item Prompt: ``nq question: who played desmond doss's father in hacksaw ridge?'' \\
              Answer: ``Hugo Weaving''
    \end{itemize}
\end{itemize}

After editing, the model should return ``Illinois Institute of Technology'' for both the edit prompt and its paraphrase, while preserving the correct answer to the neighborhood query.

\paragraph{Evaluation Metrics} We adopt three argmax-based metrics to avoid the overestimation issue common in probability-comparison methods \citep{meng2022locating,mengmass,fangalphaedit}.

\begin{description}
  \item[Efficacy (Eff.)] checks whether the edited model assigns the highest probability to the target output \(o_{\text{tar}}\) for the original query \(x\):
  \[
  o_{\text{tar}} = \arg\max_{o} P(o \mid x)
  \]

  \item[Generality (Gen.)] evaluates whether the edit generalizes to paraphrased inputs \(x_g\):
  \[
  o_{\text{tar}} = \arg\max_{o} P(o \mid x_g)
  \]

  \item[Locality (Loc.)] quantifies whether the edited model preserves performance on unrelated facts. For an unrelated query \(x_q\) with ground-truth answer \(o_q\):
  \[
  o_q = \arg\max_{o} P(o \mid x_q)
  \]
\end{description}

\begin{table}[htbp]
\centering
\caption{Mapping from energy retention ratio to effective rank for each model. Although ranks vary slightly across editing layers, they remain consistent within a narrow range.}
\label{tab:rank_energy}
\begin{tabular}{lcccccccc}
\toprule
 & \multicolumn{8}{c}{Energy Retention Ratio} \\
\cmidrule(lr){2-9}
& 0.1 & 0.3 & 0.5 & 0.7 & 0.8 & 0.9 & 0.95 & 1.0 \\
\midrule
LLaMA ($d$=14336) & 21 & 375 & 1472 & 3875 & 5958 & 9104 & 11314 & 14336 \\
GPT-J ($d$=16384)  & 1  & 9   & 639  & 3326 & 5911 & 9773 & 12453 & 16384 \\
GPT2-XL ($d$=6400)    & 1  & 95  & 629  & 1852 & 2820 & 4154 & 5047  & 6400 \\
\bottomrule
\end{tabular}
\end{table}
\textbf{Note on Locality for ZsRE} 
Unlike cloze-style tasks such as CounterFact, ZsRE is a question-answering dataset where models often fail to produce locality answers in the exact expected format, making strict token matching infeasible and locality estimates close to zero. 
To address this, we adopt the token-level autoregressive evaluation strategy from AlphaEdit, following the exact implementation in its codebase: \url{https://github.com/jianghoucheng/AlphaEdit}. 
Specifically, for a ground-truth answer \(o_q = (o_q^1, o_q^2, \dots, o_q^n)\), we sequentially feed the prefix \((o_q^1, \dots, o_q^i)\) into the model to predict the next token \(o_q^{i+1}\) and average the accuracy across all positions \(i = 0, \dots, n-1\).

\subsection{Adaptation to Sequential Editing}
\label{subapp:adapt}

To enable baseline LTE methods to scale to thousands of sequential edits, we incorporate the history-aware regularization strategy proposed by \citet{fangalphaedit}. This adaptation explicitly considers  previously edited knowledge during subsequent updates.\footnote{Without the history-aware adaptation, LTE methods typically corrupt the model within a few hundred editing steps with a batch size of 1 \citep{thedewikibigedit,liubetaedit}.}
We denote:
\begin{itemize}
    \item $\mathbf{K}_0$: key representations of a fixed set of knowledge-preserving (i.e., non-edit) samples,
    \item $\mathbf{K}_t$: keys corresponding to the current batch of edit targets at step $t$,
    \item $\mathbf{K}_{1:t-1} = [\mathbf{K}_1, \dots, \mathbf{K}_{t-1}]$: concatenated keys from all previously edited batches.
\end{itemize}

Below, we describe the adapted update rules, omitting regularization hyperparameters for clarity.

\paragraph{MEMIT:}
The original MEMIT update is:
\begin{equation}
\label{eq:upd_emmet}
    \Delta_{\text{MEMIT}} = \mathbf{R}_t \mathbf{K}_t^T \left(\mathbf{K}_0 \mathbf{K}_0^T + \mathbf{K}_t \mathbf{K}_t^T \right)^{-1}.
\end{equation}
To protect previously edited knowledge, we extend the regularization term:
\begin{equation}
    \Delta_{\text{MEMIT}} = \mathbf{R}_t \mathbf{K}_t^T \left(\mathbf{K}_0 \mathbf{K}_0^T + \mathbf{K}_{1:t-1} \mathbf{K}_{1:t-1}^T + \mathbf{K}_t \mathbf{K}_t^T \right)^{-1}.
\end{equation}
This adaptation has been applied by HSE \citep{fang2025hippocampal}. So we use HSE as the baseline.

\paragraph{EMMET:}
The original update rule is:
\begin{equation}
    \Delta_{\text{EMMET}} = \mathbf{R}_t \left(\mathbf{K}_t^T (\mathbf{K}_0 \mathbf{K}_0^T)^{-1} \mathbf{K}_t\right)^{-1} \mathbf{K}_t^T (\mathbf{K}_0 \mathbf{K}_0^T)^{-1}.
\end{equation}
We adapt it by updating the inverse covariance estimate to include prior edits:
\begin{equation}
\begin{split}
    \Delta_{\text{EMMET}} = \mathbf{R}_t \Big( & \mathbf{K}_t^T (\mathbf{K}_0 \mathbf{K}_0^T + \mathbf{K}_{1:t-1} \mathbf{K}_{1:t-1}^T)^{-1} \mathbf{K}_t \Big)^{-1}  \mathbf{K}_t^T (\mathbf{K}_0 \mathbf{K}_0^T + \mathbf{K}_{1:t-1} \mathbf{K}_{1:t-1}^T)^{-1}.
\end{split}
\end{equation}
Unlike the original EMMET, which applies no regularization coefficient to \(\mathbf{K}_0 \mathbf{K}_0^T\), we introduce a coefficient \(\lambda\) to prevent the historical terms \(\mathbf{K}_{1:t-1} \mathbf{K}_{1:t-1}^T\) from quickly overwhelming \(\mathbf{K}_0 \mathbf{K}_0^T\) and causing the model to collapse rapidly \citep{liubetaedit}. In our experiments, we set \(\lambda = 15,000\).
\paragraph{AlphaEdit:}
AlphaEdit inherently supports sequential editing by design. Its update already includes protection for previously edited knowledge:
\begin{equation}
    \Delta_{\text{AlphaEdit}} = \mathbf{R} \mathbf{K}_t^T \mathbf{P} \left(\mathbf{I} + \mathbf{K}_{1:t-1} \mathbf{K}_{1:t-1}^T \mathbf{P} + \mathbf{K}_t \mathbf{K}_t^T \mathbf{P} \right)^{-1},
\end{equation}
where $\mathbf{P}$ is a projection matrix onto the null space of preserved knowledge. Hence, no further adaptation is required. 

\subsection{Adaptation to Low-Rank Regularization}
\label{subapp:low_adapt}

\paragraph{Correspondence between rank $r$ and  energy retention ratio $\gamma$} 
Table~\ref{tab:rank_energy} summarizes the rank required to retain a given proportion of the total energy—defined via the singular value spectrum—for LLaMA3-8B, GPT-J-6B, and GPT2-XL. For each model, we compute the minimal rank such that the cumulative energy, i.e., the sum of squared singular values up to that rank divided by the total sum, reaches the specified retention ratio (e.g., 0.9 for 90\%). The reported correspondence is based on the final editing layer. Although this rank–energy mapping varies slightly across layers, the variation is minor, reflecting consistent spectral decay behavior within each model.

\paragraph{Results of knowledge perturbation on the ZsRE dataset} Figure~\ref{fig:perturb_zsre} compares the knowledge perturbation induced by HSE, AlphaEdit, and GEdit when editing the ZsRE dataset. The results demonstrate that GEdit consistently achieves zero knowledge perturbation on the core knowledge subspace $\mathbf{U}$, thereby enabling stable editing and preserving high edit locality.

\paragraph{Application to HSE, EMMET and AlphaEdit} 
The low-rank regularization is built upon the LR+D factor model, which replaces the knowledge preservation term with the expected perturbation. Thus, for MEMIT (or HSE), we have its low-rank regularized version as follows,
\begin{equation}
\label{eq:memit_gedit}
\boxed{
\Delta_{\text{MEMIT}}^{\text{GEdit}} = \mathbf{R}_1 \mathbf{K}_1^\top \big( \mathbf{K}_1 \mathbf{K}_1^\top + \beta (\mathbf{U}\mathbf{U}^\top + \bar{\mathbf{D}}) \big)^{-1}
},
\end{equation}
And for EMMET, we have its low-rank regularized version as follows,
\begin{equation}
\label{eq:emmet_gedit}
\boxed{
    \Delta_{\text{EMMET}}^{\text{GEdit}} = \mathbf{R}_1 \left(\mathbf{K}_1^\top (\beta(\mathbf{U} \mathbf{U}^\top +  \bar{\mathbf{D}}))^{-1} \mathbf{K}_1\right)^{-1} \mathbf{K}_1^\top (\beta(\mathbf{U} \mathbf{U}^\top +  \bar{\mathbf{D}}))^{-1}.}
\end{equation}
In AlphaEdit, the knowledge preservation term is canceled due to null space projection, so no additional adaptation is required except for the null space projector. 
The low-rank regularization is designed to protect the core knowledge subspace $\mathbf{U}$. Therefore, we construct the projector $\mathbf{P}$ in AlphaEdit as $\mathbf{P} = \mathbf{M}\mathbf{M}^\top$.

\begin{figure*}[t]
  \centering
  \includegraphics[width=0.95\linewidth]{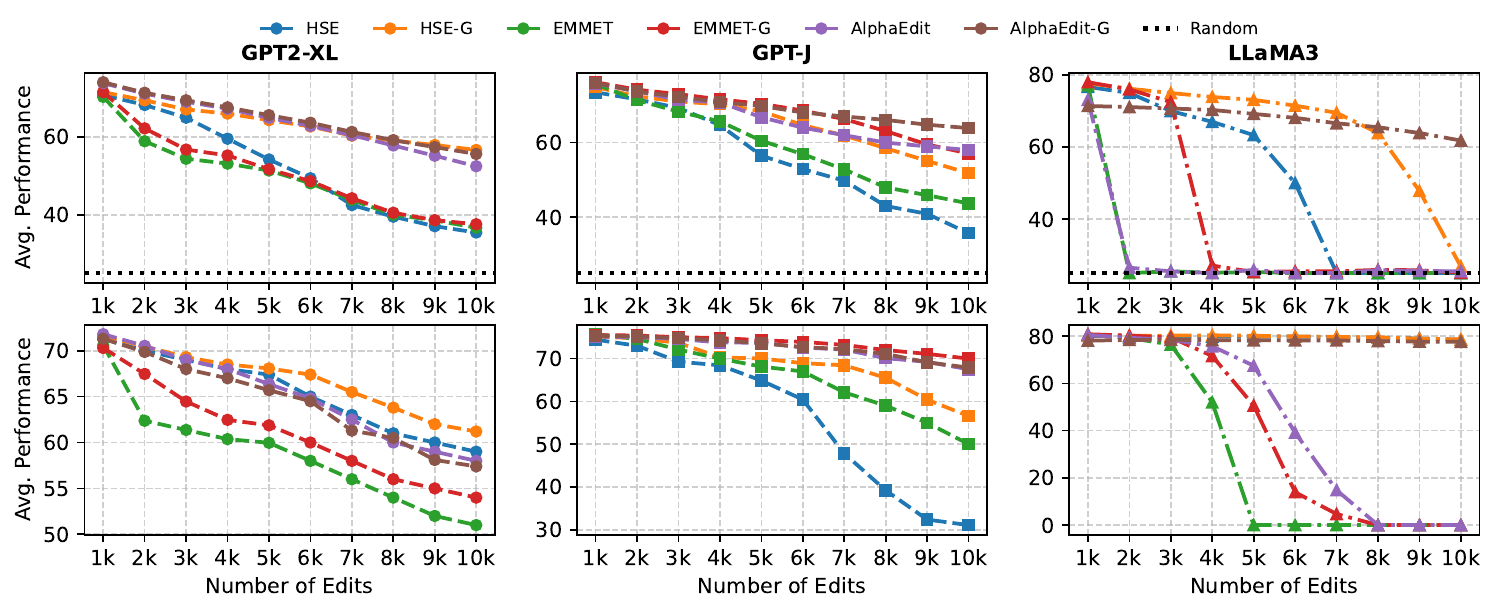}
\caption{
    Sequential editing performance on \textsc{CounterFact} (top row) and \textsc{ZsRE} (bottom row) as the number of edits scales from 1k to 10k. 
    \textsc{GEdit} variants (denoted with the ``-G'' suffix) consistently outperform their base editing methods.
}
  \label{fig:scale_raw}
\end{figure*}

With these adaptations, we compare the editing performance of low-rank regularization against full-rank regularization in Figure~\ref{fig:scale_raw}, where the low-rank version is denoted with the "-G" suffix. 
For a fair comparison, we do not tune the regularization strength separately. 
Specifically, $\beta$ is uniformly set to $15000$ for HSE, HSE-G, EMMET, and EMMET-G, while it is set to $10$ for AlphaEdit and AlphaEdit-G, following the default setting in their paper. For \textsc{CounterFact}, performance is averaged over \textbf{Eff.}, \textbf{Gen.}, \textbf{Loc.}, \textbf{Eff.*}, \textbf{Gen.*} and \textbf{Loc.*}, where the latter three are the widely adopted probability-comparison-based metrics \citep{fangalphaedit}; for \textsc{ZsRE}, it is averaged over \textbf{Eff.}, \textbf{Gen.}, and \textbf{Loc.}. 
As shown, the low-rank variants significantly outperforms their full-rank counterparts.

\begin{table}[htbp]
\centering
\caption{Performance of \textsc{GEdit} and \textsc{HSE} under fixed regularization, which requires per-method, per-dataset, and per-model tuning and is therefore suboptimal. CounterFact is abbreviated as CF.}
\label{tab:hyperparam_sensitivity}
\small
\setlength{\tabcolsep}{1.5pt}
\begin{tabular}{lccccccccccccccc}
\toprule
\multirow{2}{*}{Setting} & \multicolumn{15}{c}{Regularization strength $\mu$} \\
\cmidrule(lr){2-16}
& \multicolumn{3}{c}{5000} & \multicolumn{3}{c}{10000} & \multicolumn{3}{c}{15000} & \multicolumn{3}{c}{20000} & \multicolumn{3}{c}{50000} \\
\cmidrule(lr){2-4} \cmidrule(lr){5-7} \cmidrule(lr){8-10} \cmidrule(lr){11-13} \cmidrule(lr){14-16}
& Eff. & Gen. & Loc. & Eff. & Gen. & Loc. & Eff. & Gen. & Loc. & Eff. & Gen. & Loc. & Eff. & Gen. & Loc. \\
\midrule
GEdit / GPT-J / CF & 76.2 & 37.3 & 3.80 & 80.1 & 41.8 & 4.30 & 85.8 & 45.4 & 5.00 & 87.4 & 48.1 & 5.50 & 42.3 & 28.9 & 8.90  \\
GEdit / GPT-J / ZsRE & 69.9 & 52.4 & 20.1 & 77.2 & 69.1 & 23.8 & 86.5 & 78.3 & 25.4 & 83.0 & 74.2 & 26.3 & 45.1 & 34.5 & 27.0  \\
GEdit / LLaMA / CF & 66.1 & 41.2 & 3.90 & 72.8 & 60.9 & 5.20 & 76.3 & 61.5 & 7.40 & 71.1 & 58.1 & 9.00 & 38.6 & 27.4 & 16.7 \\
GEdit / LLaMA / ZsRE & 80.4 & 69.1 & 32.8 & 89.7 & 81.5 & 34.5 & 85.2 & 74.4 & 34.9 & 81.1 & 70.9 & 35.0 & 39.5 & 26.2 & 37.0  \\
\midrule
HSE / GPT-J / CF & 0.00 & 0.00 & 0.00 & 8.10 & 1.80 & 0.40 & 26.7 & 16.0 & 2.40 & 33.5 & 20.2 & 3.60 & 30.6 & 18.9 & 8.50  \\
HSE / GPT-J / ZsRE & 0.00 & 0.00 & 0.00 & 41.2 & 34.2 & 3.80 & 38.9 & 35.1 & 6.20 & 43.3 & 36.2 & 13.7 & 35.1 & 20.5 & 27.0  \\
HSE / LLaMA / CF & 0.00 & 0.00 & 0.00 & 0.00 & 0.00 & 0.00 & 0.00 & 0.00 & 0.00 & 0.00 & 0.00 & 0.00 & 40.2 & 29.5 & 16.4 \\
HSE / LLaMA / ZsRE & 0.00 & 0.00 & 0.00 & 81.7 & 70.5 & 33.5 & 90.2 & 82.1 & 35.2 & 85.3 & 75.6 & 35.8 & 37.2 & 29.2 & 36.9  \\
\bottomrule
\end{tabular}
\end{table}

\subsection{Adaptation to Adaptive Regularization}
\label{subapp:ada_adapt}
\paragraph{Comparison with fixed regularization} We provide a detailed comparison between our adaptive regularization framework and fixed regularization approaches. Table~\ref{tab:hyperparam_sensitivity} presents the editing performance of two editors (\textsc{GEdit} and \textsc{HSE}) under fixed regularization across varying regularization strengths $\mu \in \{5000, 10000, 15000, 20000, 50000\}$, evaluated on two models (GPT-J and LLaMA) and two datasets (CounterFact and ZsRE).

A key observation from Table~\ref{tab:hyperparam_sensitivity} is that there is no single fixed regularization strength that works optimally across different combinations of methods, models, and datasets. For example, for \textsc{GEdit} with GPT-J on CounterFact, the optimal $\mu$ is around 20000, achieving 87.4\% efficacy, 48.1\% generalization, and 5.50 locality. However, for \textsc{GEdit} with LLaMA on ZsRE, the best efficacy (89.7\%) occurs at $\mu = 10000$. For \textsc{HSE} with GPT-J on CounterFact, meaningful performance only emerges at $\mu \ge 10000$, with optimal trade-offs around $\mu = 20000$. More dramatically, \textsc{HSE} with LLaMA on CounterFact produces all-zero metrics (complete failure) for $\mu \le 20000$, only achieving reasonable performance at $\mu = 50000$.

These results clearly demonstrate that the optimal fixed regularization strength is highly sensitive to the choice of editing method, base model, and evaluation dataset. A setting that performs well in one configuration may lead to suboptimal or even completely failed edits in another.

Comparing Table~\ref{tab:hyperparam_sensitivity} with the main text's Tables \ref{tab:results} and \ref{tab:ada_r} reveals a consistent pattern: adaptive regularization substantially outperforms even the best-tuned fixed regularization across all configurations. More importantly, the adaptive approach achieves this without requiring per-method, per-dataset, per-model hyperparameter tuning. This is because the correspondence between editability and output perturbation (Theorem \ref{th:ee}) is agnostic to these factors.

\begin{figure}[htbp]
  \centering
  \begin{minipage}[t]{0.49\linewidth}
    \centering
    \includegraphics[width=\linewidth]{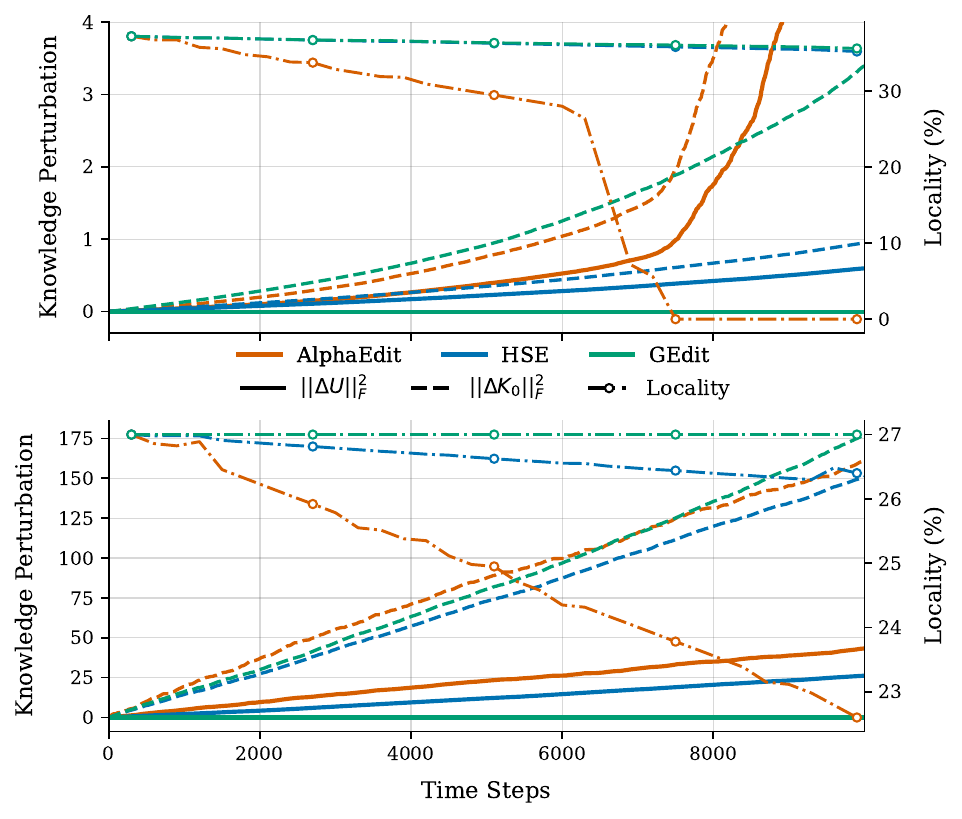}
    \caption{Evolution of knowledge perturbation (left axis) and editing locality (right axis) during 10,000 sequential edits of LLaMA (top) and GPT-J (bottom).}
    \label{fig:perturb_zsre}
  \end{minipage}
  \hfill 
  \begin{minipage}[t]{0.49\linewidth}
    \centering
    \includegraphics[width=\linewidth]{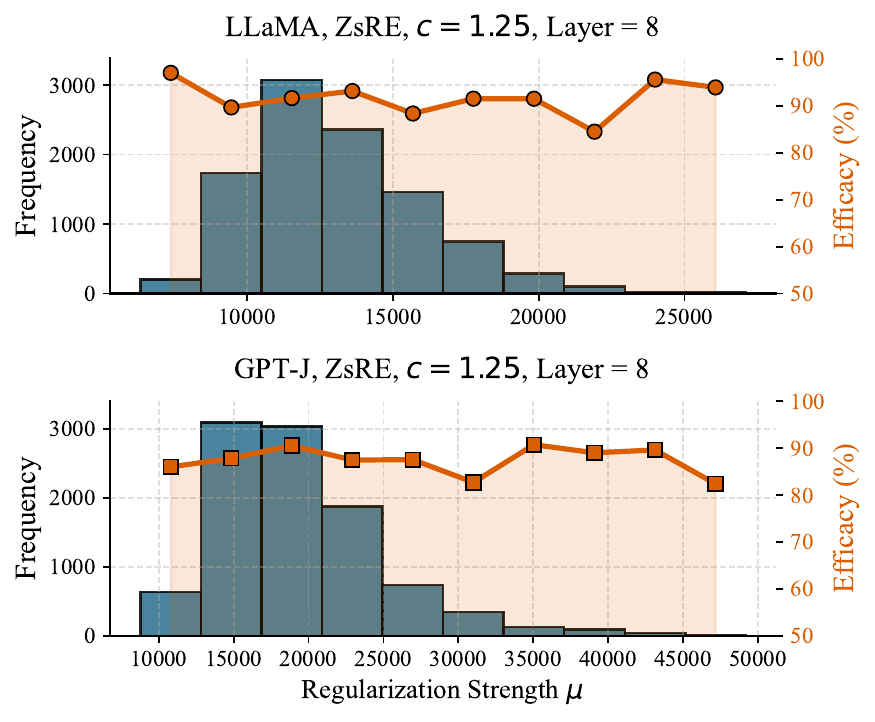}
    \caption{Frequency distribution of the adaptive regularization strength across 10,000 sequential edits, along with the corresponding edit accuracy per strength bin.}
    \label{fig:mu_zsre}
  \end{minipage}
\end{figure}

\paragraph{Results of adaptive regularization on ZsRE} 
Figure~\ref{fig:mu_zsre} shows the frequency distribution of the adaptive regularization strength across 10,000 sequential editing steps (batch size = 1) on the \textsc{ZsRE} dataset. As illustrated, the adaptive regularization mechanism dynamically adjusts the regularization strength per edit: it relaxes constraints for difficult edits to improve success rates, while tightening them for easier edits to enhance locality. Together with the results on \textsc{CounterFact} (Figure~\ref{fig:mu}), where the distribution of regularization strengths also varies, these findings demonstrate the effectiveness of our adaptive regularization strategy—capable of adjusting not only across edits but also across datasets and models.

\paragraph{Application to MEMIT (HSE).}

For MEMIT, we define editability as (Eq. \ref{eq:editability_c})
\begin{equation}
\label{eq:editability_c1}
    \mathcal{E}_{\text{MEMIT}}(\mathbf{k}_e) = \frac{1}{\mu} \mathbf{k}_e^\top (\mathbf{K}_0 \mathbf{K}_0^\top)^{-1} \mathbf{k}_e.
\end{equation}
Intuitively, this definition reflects two factors. 
A larger regularization strength \(\mu\) penalizes model changes more heavily, reducing editability — hence \(\mu\) appears in the denominator. 
The quadratic form \(\mathbf{k}_e^\top (\mathbf{K}_0 \mathbf{K}_0^\top)^{-1} \mathbf{k}_e\) (the squared Mahalanobis distance (see Appendix \ref{subapp:mahal_derivation})) measures how far \(\mathbf{k}_e\) deviates from the pretrained knowledge distribution: 
if \(\mathbf{k}_e\) lies in a low-variance direction of \(\mathbf{K}_0 \mathbf{K}_0^\top\), this term is large, 
indicating that the pre-trained model has weak knowledge there and should be more willing to adapt, 
thus increasing editability. 

The MEMIT update rule for an edit \((\mathbf{k}_e,\mathbf{r}_e)\) is given by
\begin{equation}
    \Delta_{\text{MEMIT}} = \mathbf{r}_e \mathbf{k}_e^\top \left( \mu \mathbf{K}_0 \mathbf{K}_0^\top + \mathbf{K}_e \mathbf{K}_e^\top \right)^{-1}.
\end{equation}

We now formally establish the relationship between editability and editing success: 
higher editability leads to greater success, as stated in the following theorem.

\begin{theorem}
Let \((\mathbf{k}_e, \mathbf{v}_e)\) be a knowledge pair to be edited, and let 
\(\mathbf{r}_e = \mathbf{v}_e - \mathbf{W}\mathbf{k}_e\) be the corresponding output residual. 
Denote by \(\mathcal{E}_{\text{MEMIT}}(\mathbf{k}_e)\) the editability defined in Eq.~\eqref{eq:editability_c1}. 
Then the output perturbation induced by the MEMIT update on \(\mathbf{k}_e\) satisfies
\[
    \Delta_{\text{MEMIT}} \mathbf{k}_e = \frac{\mathcal{E}_{\text{MEMIT}}(\mathbf{k}_e)}{1 + \mathcal{E}_{\text{MEMIT}}(\mathbf{k}_e)} \, \mathbf{r}_e.
\]
\end{theorem}

Thus, a larger editability \(\mathcal{E}_{\text{MEMIT}}(\mathbf{k}_e)\) makes the factor 
\(\frac{\mathcal{E}}{1+\mathcal{E}}\) closer to \(1\), meaning the update nearly recovers 
the full residual \(\mathbf{r}_e\) and achieves near-perfect editing success.

By setting \(\mathcal{E}_{\text{MEMIT}}(\mathbf{k}_e) = c\) with a constant hyperparameter \(c > 0\), 
we obtain an edit-adaptive regularization strength
\begin{equation}
    \mu = \frac{1}{c} \, \mathbf{k}_e^\top (\mathbf{K}_0 \mathbf{K}_0^\top)^{-1} \mathbf{k}_e.
\end{equation}
\paragraph{Application to EMMET (ROME).}

For EMMET, the update rule for an edit \((\mathbf{k}_e,\mathbf{r}_e)\) is given by
\begin{equation}
\label{eq:emmet_update}
    \Delta_{\text{EMMET}} = \mathbf{r}_e \left( \mathbf{k}_e^\top (\mu \mathbf{K}_0 \mathbf{K}_0^\top)^{-1} \mathbf{k}_e + a I \right)^{-1} \mathbf{k}_e^\top (\mu \mathbf{K}_0 \mathbf{K}_0^\top)^{-1},
\end{equation}
where $\mu > 0$ and $a > 0$ are regularization hyperparameters, with $a$ not only protecting knowledge but also controlling editing stability \citep{meng2022locating,gupta2024unified}.

We define editability for EMMET as
\begin{equation}
\label{eq:editability_emmet}
    \mathcal{E}_{\text{EMMET}}(\mathbf{k}_e) = \frac{1}{\mu a} \, \mathbf{k}_e^\top (\mathbf{K}_0 \mathbf{K}_0^\top)^{-1} \mathbf{k}_e.
\end{equation}
Intuitively, 
A larger regularization strength \(\mu\) or \(a\) penalizes model changes more heavily, reducing editability — hence the product \(\mu a\) appears in the denominator. 

We now formally establish the relationship between editability and editing success: 
higher editability leads to greater success, as stated in the following theorem.

\begin{theorem}
Let \((\mathbf{k}_e, \mathbf{v}_e)\) be a knowledge pair to be edited, and let 
\(\mathbf{r}_e = \mathbf{v}_e - \mathbf{W}\mathbf{k}_e\) be the corresponding output residual. 
Denote by \(\mathcal{E}_{\text{EMMET}}(\mathbf{k}_e)\) the editability defined in Eq.~\eqref{eq:editability_emmet}. 
Then the output perturbation induced by the EMMET update on \(\mathbf{k}_e\) satisfies
\[
    \Delta_{\text{EMMET}} \mathbf{k}_e = \frac{\mathcal{E}_{\text{EMMET}}(\mathbf{k}_e)}{1 + \mathcal{E}_{\text{EMMET}}(\mathbf{k}_e)} \, \mathbf{r}_e.
\]
\end{theorem}

Thus, a larger editability \(\mathcal{E}_{\text{EMMET}}(\mathbf{k}_e)\) makes the factor 
\(\frac{\mathcal{E}}{1+\mathcal{E}}\) closer to \(1\), meaning the update nearly recovers 
the full residual \(\mathbf{r}_e\) and achieves near-perfect editing success.

By setting \(\mathcal{E}_{\text{EMMET}}(\mathbf{k}_e) = c\) with a constant hyperparameter \(c > 0\), 
we obtain an edit-adaptive regularization condition
\begin{equation}
\label{eq:adaptive_condition_emmet}
    \mu = \frac{1}{ca} \, \mathbf{k}_e^\top (\mathbf{K}_0 \mathbf{K}_0^\top)^{-1} \mathbf{k}_e.
\end{equation}
Or, we can fix $\mu$ and tune $a$. This provides a flexible two-parameter adaptive regularization strategy.

\paragraph{Application to AlphaEdit.}

For AlphaEdit, we define editability analogously. 
Under the projection-based update, editability is formulated as
\begin{equation}
\label{eq:editability_alpha}
    \mathcal{E}_{\text{Alpha}}(\mathbf{k}_e) = \frac{1}{\mu} \mathbf{k}_e^\top \mathbf{P} \mathbf{k}_e,
\end{equation}
where \(\mathbf{P}\) is a projection matrix satisfying \(\mathbf{P}^2 = \mathbf{P}\) and \(\mathbf{P}^\top = \mathbf{P}\). Intuitively, \(\mathbf{k}_e^\top \mathbf{P} \mathbf{k}_e = \|\mathbf{P} \mathbf{k}_e\|^2\) measures the component of \(\mathbf{k}_e\) 
that lies in the null space where editing is permitted. 
If \(\|\mathbf{P} \mathbf{k}_e\|\) is small, most of \(\mathbf{k}_e\) does not fall into the null space, 
meaning the edit direction is ignored by the projection and thus harder to effect — hence editability is low. 
Conversely, a large \(\|\mathbf{P} \mathbf{k}_e\|\) indicates that \(\mathbf{k}_e\) is well-aligned with the null space, 
making the edit easier. The regularization strength \(\mu\) again appears in the denominator, 
as stronger regularization reduces editability.

The AlphaEdit update rule for an edit \((\mathbf{k}_e,\mathbf{r}_e)\) is given by
\begin{equation}
    \Delta_{\text{AlphaEdit}} = \mathbf{r}_e \mathbf{k}_e^\top \mathbf{P} \left( \mathbf{k}_e \mathbf{k}_e^\top \mathbf{P} + \mu I \right)^{-1}.
\end{equation}

We now establish the analogous relationship for AlphaEdit.

\begin{theorem}
Let \((\mathbf{k}_e, \mathbf{v}_e)\) be a knowledge pair to be edited, and let 
\(\mathbf{r}_e = \mathbf{v}_e - \mathbf{W}\mathbf{k}_e\) be the corresponding output residual. 
Let \(\mathbf{P}\) be a projection matrix as defined above, and let 
\(\mathcal{E}_{\text{Alpha}}(\mathbf{k}_e)\) be the editability defined in Eq.~\eqref{eq:editability_alpha}. 
Then the output perturbation induced by the AlphaEdit update on \(\mathbf{k}_e\) satisfies
\[
    \Delta_{\text{AlphaEdit}} \mathbf{k}_e = \frac{\mathcal{E}_{\text{Alpha}}(\mathbf{k}_e)}{1 + \mathcal{E}_{\text{Alpha}}(\mathbf{k}_e)} \, \mathbf{r}_e.
\]
\end{theorem}

As before, a higher editability brings the factor \(\frac{\mathcal{E}}{1+\mathcal{E}}\) closer to \(1\), 
enabling the update to nearly recover the desired residual \(\mathbf{r}_e\) and achieve successful editing.

By setting \(\mathcal{E}_{\text{Alpha}}(\mathbf{k}_e) = c\) with the same hyperparameter \(c > 0\) 
(e.g., \(c = 1.25\)), we obtain an edit-adaptive regularization strength for AlphaEdit:
\begin{equation}
    \mu = \frac{1}{c} \, \mathbf{k}_e^\top \mathbf{P} \mathbf{k}_e.
\end{equation}
\begin{figure*}[t]
  \centering
  \includegraphics[width=0.95\linewidth]{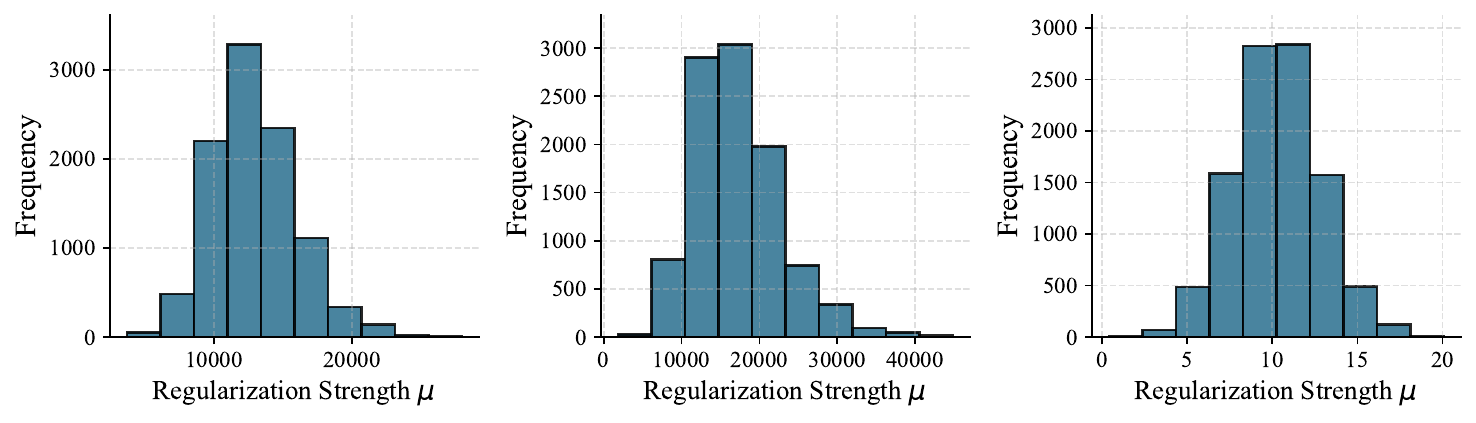}
  \caption{
    Adaptive regularization strength across 10,000 edits on the CounterFact dataset, applied to the LLaMA model. The desired editability is set to $1.25$. From left to right: EMMET, MEMIT, AlphaEdit.
  }
  \label{fig:mu_other_methods}
\end{figure*}

\begin{figure*}[t]
  \centering
  \includegraphics[width=0.95\linewidth]{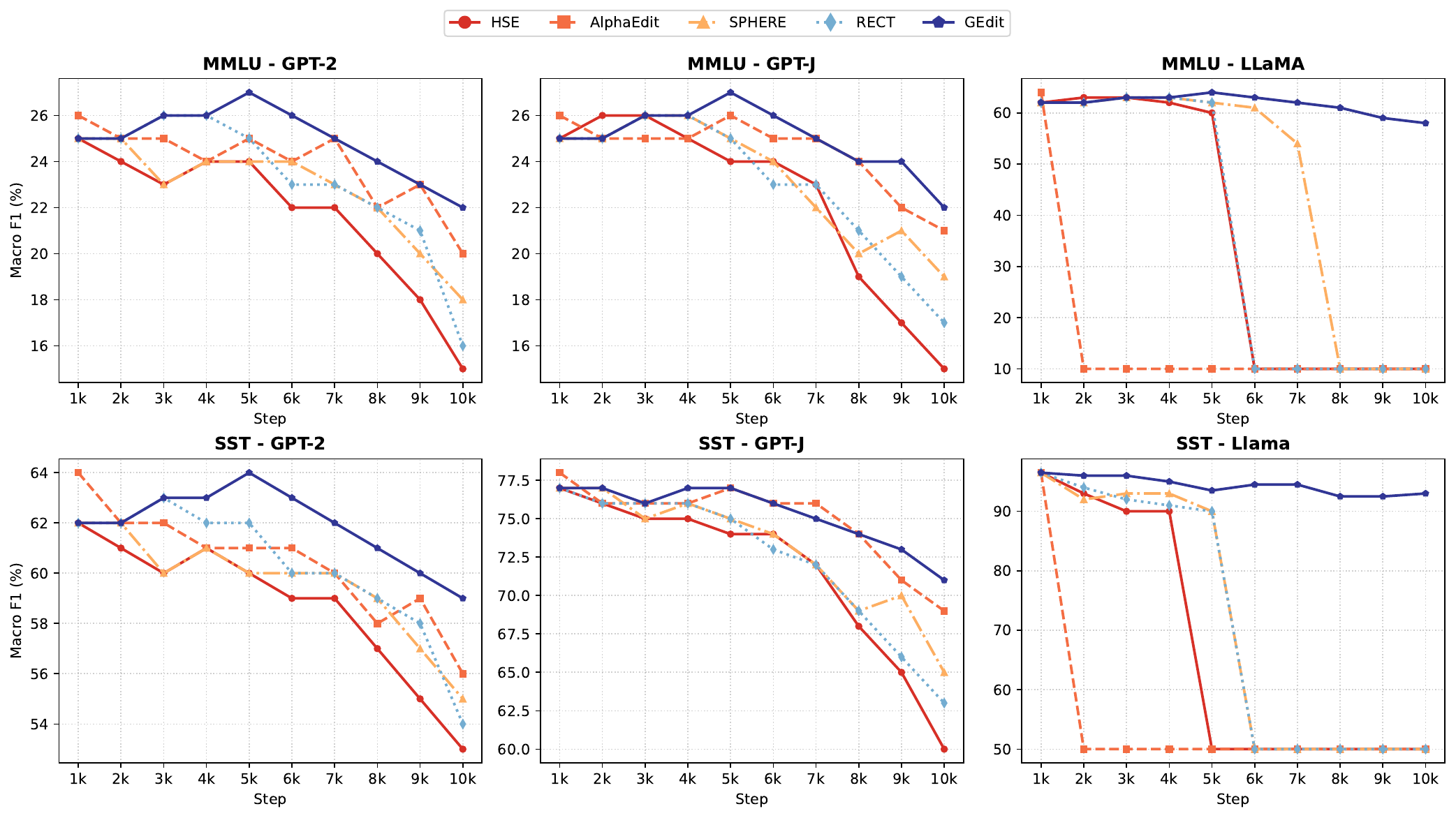}
\caption{
    SST and MMLU performance across 10,000 sequential edits on the CounterFact dataset, evaluated every 1,000 steps (10 evaluation points total). Results are shown for three base models (GPT-2, GPT-J, and LLaMA) across five editing methods: HSE, AlphaEdit, SPHERE, RECT, and GEdit.
}
  \label{fig:sst_mmlu}
\end{figure*}

\begin{table}[htbp]
\centering
\caption{Editing performance under 2,000 sequential edits (batch size=1).}
\label{tab:results_2000}
\footnotesize
\setlength{\tabcolsep}{3pt}
\begin{tabular}{l l c c c c c c c c c c c}
\toprule
\multirow{2}{*}{Method} & & \multicolumn{7}{c}{CounterFact} & \multicolumn{4}{c}{ZsRE} \\
\cmidrule(lr){3-9} \cmidrule(lr){10-13}
& & SST& Eff*. & Gen*. & Loc*. & Eff. & Gen. & Loc. & SST& Eff. & Gen. & Loc. \\
\midrule
Pre-edited&& 62.5&22.1&24.4&78.0&0.10&0.40&10.6& 62.5&0.00&0.00&25.0\\
\midrule
HSE &\multirow{4}{*}{\rotatebox{90}{GPT2-XL}}& 62.0 &98.0& 88.6& 65.7&91.4&58.4 &10.6& 60.0&94.7&88.4&26.9  \\
EMMET    && 62.0&92.4 & 85.6& 57.1&71.1&49.9 &5.50& 62.0&85.0 &77.8&24.3  \\
AlphaEdit && 62.5& 99.6&94.0 & 65.7&97.3&65.3 &6.40& 62.5& 96.0&88.8& 26.8\\
GEdit && 62.5& 99.6&94.0 & 65.6&96.8&65.6 &6.40& 62.5& 95.2&88.0& 26.4\\
\midrule
Pre-edited&&77.0&15.4&18.0&83.3&0.40&0.60&13.7&77.0&0.00&0.00&27.4\\
\midrule

MEMIT  & \multirow{4}{*}{\rotatebox{90}{GPT-J}} &77.5& 99.2&89.9 & 77.5&96.4& 51.3&9.90&72.0&98.8&93.3& 27.5 \\
EMMET   & &77.0&99.6 & 94.4& 75.3&98.1& 60.6&10.4&75.0&99.8 &97.2& 28.6 \\
AlphaEdit  &&78.0&99.8 & 93.8&76.6 &99.0& 61.2&9.90&72.5&99.8 &97.8&28.7 \\
GEdit& &77.5&99.8 & 94.6&77.8 &99.4& 65.4&10.8&76.0&99.8 &97.9&28.9 \\
\midrule
Pre-edited&&96.5&7.80&10.4&89.3&0.30&0.50&21.3&96.5&0.00&0.00&38.6\\
\midrule


HSE  & \multirow{4}{*}{\rotatebox{90}{LLaMA3}} &94.5&99.0 &92.9 &71.5 & 97.0&71.3 &18.4&97.0&99.2&95.4&38.9 \\
EMMET      & &50.0 &50.1 & 50.0&50.0 &0.00&0.00 &0.00&93.0&97.3 &93.9&38.0 \\
AlphaEdit   & & 50.5 &50.1 & 50.0&50.5 &0.00& 0.00&0.60 &95.5&98.0 &94.2&38.4 \\
GEdit  &    &96.0 & 99.6& 93.0&78.1 &98.4&68.4 &20.6 &97.0&99.5&96.0&38.7  \\
\bottomrule
\end{tabular}%
\normalsize
\end{table}

In Figure~\ref{fig:mu_other_methods}, we demonstrate that although different methods employ varying regularization strengths, they can be uniformly modulated through the concept of editability. Specifically, we set the desired editability to $1.25$ for EMMET, MEMIT, and AlphaEdit. Under this unified configuration, the regularization strength of each method is automatically tuned across edits, as shown in the figure. This explains why AlphaEdit originally sets its default regularization strength $\mu$ to $10$—it corresponds approximately to the average value of our adaptive regularization strength. The same observation applies to EMMET and MEMIT.
\paragraph{Adaptation to Sequential Editing}
We extend the adaptive regularization mechanism of GEdit to sequential
editing, where edits are applied over multiple steps while preserving both
pretrained knowledge and previously edited knowledge. At step $t$, let
$\mathbf{K}_t$ and $\mathbf{V}_t$ denote the keys and desired values of the
current edit batch, respectively, and let
$\mathbf{K}_{1:t-1}=[\mathbf{K}_1,\ldots,\mathbf{K}_{t-1}]$ contain the keys
of all previously edited batches. The current model weights are denoted by
$\mathbf{W}_{t-1}$, and the target residual is
$\mathbf{R}_t=\mathbf{V}_t-\mathbf{W}_{t-1}\mathbf{K}_t$. To preserve both
pretrained and previously edited knowledge, we introduce the unified
regularization matrix \citep{fangalphaedit,fang2025hippocampal,guotowards,liubetaedit}
\begin{equation}
\widetilde{\mathbf{D}}_t
=
\mathbf{K}_{1:t-1}\mathbf{K}_{1:t-1}^\top
+\mu\bar{\mathbf{D}},
\label{eq:unified-D}
\end{equation}
where $\mu>0$ is a fixed hyperparameter controlling the relative contribution
of pretrained knowledge preservation. Introducing an adaptive coefficient
$\lambda_t>0$, we obtain the following weight update for sequential editing:
\begin{equation}
\Delta_t
=
\mathbf{R}_t\mathbf{K}_t^\top\mathbf{M}
\left[
\mathbf{M}^\top
\left(
\mathbf{K}_t\mathbf{K}_t^\top
+\lambda_t\widetilde{\mathbf{D}}_t
\right)
\mathbf{M}
\right]^{-1}
\mathbf{M}^\top.
\label{eq:closed-form}
\end{equation}

We can correspondingly define the history-aware
editability as
\begin{equation}
\mathcal{E}_t(\mathbf{k}_t)
=
\frac{1}{\lambda_t}
\mathbf{k}_t^\top\mathbf{M}
\left(
\mathbf{M}^\top\widetilde{\mathbf{D}}_t\mathbf{M}
\right)^{-1}
\mathbf{M}^\top\mathbf{k}_t.
\label{eq:history-editability}
\end{equation}
This definition generalizes the subspace-aware editability in
Eq.~\eqref{eq:editability_d} by incorporating the covariance-like structure of
previously edited keys. In particular, the term
$\mathbf{K}_{1:t-1}\mathbf{K}_{1:t-1}^\top$ penalizes updates along directions
associated with historical edits, while $\mu\bar{\mathbf{D}}$ regularizes
changes to pretrained knowledge.

Following the same argument as in Theorem~\ref{th:ee}, one can show that the
relation between editability and output residual fitting still holds in the
sequential setting, i.e.,
\begin{equation}
\Delta_t\mathbf{k}_t
=
\frac{\mathcal{E}_t(\mathbf{k}_t)}
{1+\mathcal{E}_t(\mathbf{k}_t)}
\mathbf{r}_t,
\label{eq:residual-fitting}
\end{equation}
Consequently, the adaptive coefficient $\lambda_t$ can be determined in the
same manner, i.e., by setting the desired editability to $c$, so that
$\mathcal{E}_t(\mathbf{k}_t)=c$, and then solving for $\lambda_t$.

At the first editing step, $\mathbf{K}_{1:0}$ is empty, and the unified matrix
reduces to $\widetilde{\mathbf{D}}_1=\mu\bar{\mathbf{D}}$. Consequently, the
formulation recovers the original subspace-aware regularization structure.

\subsection{Additional Experimental Results}
\label{app:results}

\paragraph{Results on General-Capability Benchmarks}
Following prior work~\citep{fangalphaedit}, we evaluate the model's general capabilities on two representative benchmarks, the Stanford Sentiment Treebank (SST)~\citep{socher2013recursive} and MMLU~\citep{hendrycksmeasuring}, to assess whether its performance degrades after editing. This evaluation provides an additional measure of knowledge preservation. SST is a binary sentiment classification dataset comprising sentences from movie reviews, where each input is classified as either positive (label = 1) or negative (label = 0). For example, the sentence \textit{``What better message than `love thyself' could young women of any size receive?''} is annotated with label 1 (positive). Performance on this task is measured by accuracy, denoted by \textbf{SST}. We evaluate the model in a zero-shot prompting setting using the prompt template \texttt{Review:\{input\}\textbackslash nSentiment:}, with predictions determined by comparing the probabilities of the tokens ``positive'' and ``negative''. The evaluation is conducted on 200 SST samples.

MMLU comprises multiple-choice questions spanning 57 diverse subjects, including humanities, STEM, and social sciences. Each question has four answer choices, with exactly one correct answer. For example, a typical question from the \textit{abstract algebra} category is: \textit{``Let \( G \) be a finite group with an even number of elements. Then which of the following must be true? (A) \( G \) has an element of order 2 (B) \( G \) has no element of order 2 (C) \( G \) is cyclic (D) \( G \) is abelian.''} The correct answer is (A). We adopt a zero-shot prompting strategy, using the template \texttt{Question:\{input\}\textbackslash nOptions:\{choices\}\textbackslash nAnswer:}, and compare the model's predicted token probabilities against the canonical answer letters (A/B/C/D). The evaluation is conducted on 200 MMLU samples. 
\begin{table}[htbp]
\centering
\caption{Editing performance under 5,000 sequential edits (batch size=1).}
\label{tab:editing_5000}

\footnotesize
\setlength{\tabcolsep}{3pt}
\begin{tabular}{l l c c c c c c c c c c c}
\toprule
\multirow{2}{*}{Method} & & \multicolumn{7}{c}{CounterFact} & \multicolumn{4}{c}{ZsRE} \\
\cmidrule(lr){3-9} \cmidrule(lr){10-13}
& & SST& Eff*. & Gen*. & Loc*. & Eff. & Gen. & Loc. &SST& Eff. & Gen. & Loc. \\
\midrule
Pre-edited&&62.5&21.4&23.8&78.2&0.20&0.40&10.6&62.5&0.00&0.00&24.3\\
\midrule
HSE &\multirow{4}{*}{\rotatebox{90}{GPT2-XL}}&  60.0&89.6& 76.2& 58.0&60.9&34.1 &6.30& 61.0&92.3&84.6&25.3  \\
EMMET    &&58.5&87.4 & 76.3& 55.5&54.6&30.6 &3.70& 61.5&82.1 &76.0&21.8  \\
AlphaEdit &&  62.0&98.0& 88.3 & 60.8&88.7& 52.9 &4.70& 62.0& 91.2&84.3& 23.6\\
GEdit && 62.0&97.8&87.7 & 61.0&85.4&50.1 &4.90& 62.0& 90.6&83.3& 23.2\\
\midrule
Pre-edited&&77.0&14.7&17.2&83.5&0.40&0.50&14.3&77.0&0.00&0.00&27.0\\
\midrule
HSE &\multirow{4}{*}{\rotatebox{90}{GPT-J}}  &75.0& 95.8& 86.0& 66.1&71.5&37.7 &6.60&67.0 &86.8&80.4&23.1  \\
EMMET & &73.5 & 92.0& 84.3& 67.2&71.5&41.3 &6.80& 79.0&97.3&93.1&26.0  \\
AlphaEdit &  & 76.0& 99.3& 90.2&71.9&95.7&51.9 &7.10& 76.0&99.1&95.0&27.0  \\
GEdit &  &77.0 &99.6 &91.8 & 72.4&98.1& 57.9&8.20&79.0 &99.6&96.6& 27.3\\
\midrule

Pre-edited&&96.5&7.00&9.60&89.6&0.30&0.30&21.6&96.5&0.00&0.00&38.5\\
\midrule
HSE &\multirow{4}{*}{\rotatebox{90}{LLaMA3}} &54.0& 96.8& 89.2& 61.8&87.1&65.4 &14.4&95.5& 99.0&95.1&39.5  \\
EMMET &&50.0& 51.1 & 50.7 & 50.0 & 0.00 & 0.00 & 0.00 &50.5 &1.10 & 1.10 & 0.80 \\
AlphaEdit &&50.5& 50.0 & 50.0 & 50.0 & 0.00 & 0.00 & 0.00 & 90.5&84.2 & 79.0 & 39.3 \\
GEdit &&94.0& 99.3 & 92.3 & 66.5 & 96.4 & 67.8 & 15.0 & 96.0&99.1 & 95.2 & 39.0 \\
\bottomrule
\end{tabular}%
\normalsize
\end{table}

Figure~\ref{fig:sst_mmlu} presents the SST (sentiment analysis) and MMLU (massive multitask language understanding) performance across 10,000 sequential edits on the CounterFact dataset, with evaluations conducted every 1,000 steps. Across all three base models—GPT-2, GPT-J, and LLaMA—we observe that most editing methods suffer from significant performance degradation as the number of sequential edits increases. Specifically, HSE, AlphaEdit, SPHERE, and RECT exhibit sharp drops in both SST and MMLU accuracy after a certain number of edits, often falling to near-random performance levels. In contrast, GEdit maintains remarkably stable performance throughout the entire editing sequence. For instance, on LLaMA, where the baseline MMLU Macro F1 score is 62\%, GEdit achieves 64\% at step 1,000 and only gradually declines to 58\% by step 10,000, while other methods collapse to 10\% within the first few thousand edits. Similarly, on SST with LLaMA (baseline 96.5\%), GEdit sustains accuracy above 92\% across all 10,000 edits, whereas competing methods drop to 50\% and remain there. These results demonstrate that GEdit is uniquely capable of sequential model editing without catastrophic forgetting, preserving the model's original language understanding capabilities even after extensive modifications.

\begin{figure*}[t]
  \centering
  \includegraphics[width=0.95\linewidth]{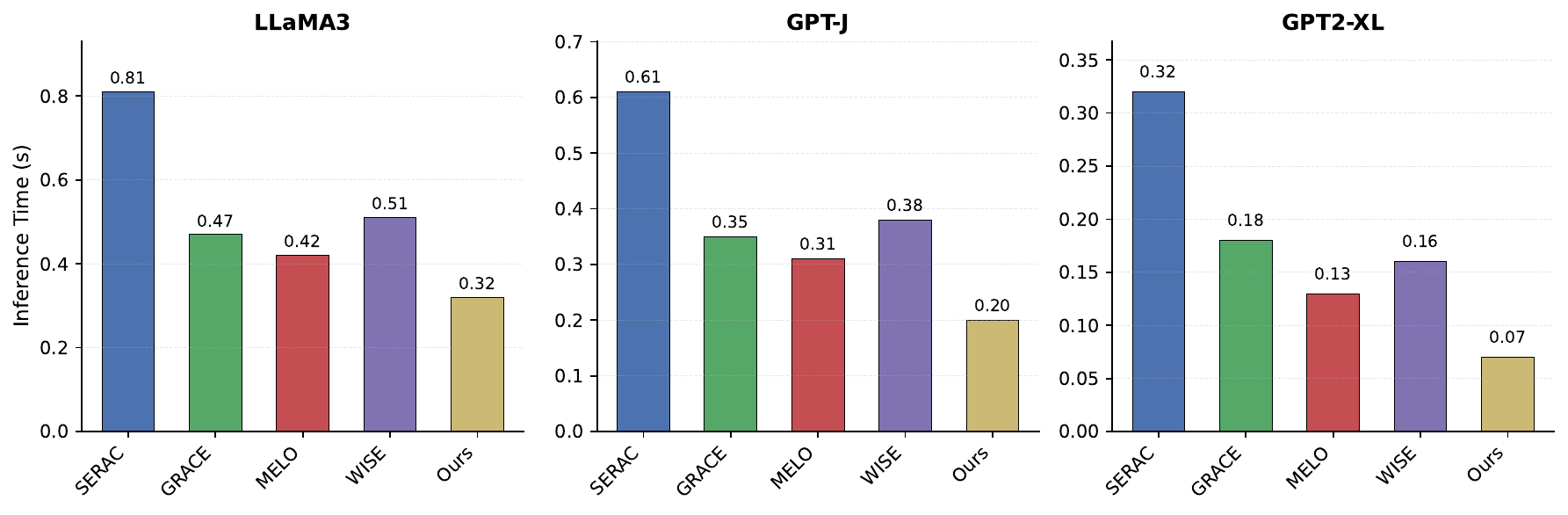}
\caption{
    Inference time (in seconds) per input prompt on LLaMA3, GPT-J, and GPT2-XL. Our method consistently achieves the lowest latency across all backbone models.
}
  \label{fig:inference_time}
\end{figure*}

\begin{table}[t]
\centering
\caption{Performance comparison (in percentage) across 10,000 independent edits.}
\label{tab:precise_edit_results}
\begin{tabular}{l|ccc|ccc|ccc}
\hline
\multirow{2}{*}{Method} & \multicolumn{3}{c|}{GPT2-XL} & \multicolumn{3}{c|}{GPT-J} & \multicolumn{3}{c}{LLaMA3-8B} \\
\cline{2-10}
& Eff. & Loc. & CP. & Eff. & Loc. & CP. & Eff. & Loc. & CP. \\
\hline
ROME & 99.7 & 10.4 & 4.8 & 99.9 & 13.8 & 5.5 & 99.9 & 21.2 & 5.3 \\
MEMIT & 89.4 & 10.6 & 2.6 & 99.7 & 14.0 & 3.2 & 96.3 & 21.6 & 3.3 \\
AlphaEdit & 98.4 & 10.5 & 3.4 & 99.7 & 14.0 & 3.0 & 96.5 & 21.7 & 3.0 \\
\textsc{GEdit} & 99.2 & 10.6 & 1.0 & 99.9 & 14.0 & 1.2 & 99.9 & 21.6 & 1.0 \\
\hline
\end{tabular}
\end{table}

\begin{table}[htbp]
\centering
\caption{
Performance comparison with parameter-preserving editing methods across three base models (GPT2-XL, GPT-J, and LLaMA3) on the CounterFact and ZsRE datasets under 2,000 sequential edits (batch size=1).
}
\label{tab:editing_2000_memo}

\footnotesize
\setlength{\tabcolsep}{5pt}
\begin{tabular}{l l c c c c c c c c c}
\toprule
\multirow{2}{*}{\textbf{Method}} & & \multicolumn{6}{c}{\textbf{CounterFact}} & \multicolumn{3}{c}{\textbf{ZsRE}} \\
\cmidrule(lr){3-8} \cmidrule(lr){9-11}
& & Eff*. & Gen*. & Loc*. & Eff. & Gen. & Loc.& Eff. & Gen. & Loc. \\
\midrule
Pre-edited&&22.1&24.4&78.0&0.10&0.40&10.6&0.0&0.0&25.0\\
\midrule
SERAC     &\multirow{4}{*}{\rotatebox{90}{GPT2-XL}}& 65.2&48.8 & 66.4&61.8&19.1 &10.1&72.7 &38.2& 26.7 \\
GRACE     && 98.8&53.4 & 73.9&96.7&21.8 &10.4&94.8 &26.9& 25.5 \\
MELO     && 95.2&66.3 & 63.5&91.9&41.6&9.80&93.5 &65.2& 27.2 \\
WISE      && 94.0&72.4 & 70.0&89.5&49.4&10.2 &91.0 &72.4& 25.9 \\

\midrule
Pre-edited&&15.4&18.0&83.3&0.40&0.60&13.7&0.0&0.0&27.4\\
\midrule

SERAC     &\multirow{4}{*}{\rotatebox{90}{GPT-J}}& 75.5 & 49.4 & 69.1 & 71.5 & 35.4 & 12.1 & 72.0 & 38.7 & 26.5 \\
GRACE     && 98.5 & 55.6 & 74.0 &95.0 & 22.5 & 13.1 & 96.8 & 31.5 & 27.0 \\
MELO      && 94.9 & 76.8 & 67.0 & 91.7 & 42.4 & 11.1 & 93.2 & 42.3 & 25.8 \\
WISE      && 96.0 & 81.2 & 72.0 & 93.6 & 61.6 & 12.5 & 97.6 & 73.4 & 28.0 \\

\midrule
Pre-edited&&7.80&10.4&89.3&0.30&0.50&21.3 &0.0&0.0&38.6\\
\midrule
SERAC     &\multirow{4}{*}{\rotatebox{90}{LLaMA3}}& 63.4&53.5 & 69.0&60.2&18.4 &19.1&60.5& 37.8 &36.7 \\
GRACE     && 97.0&52.3 & 79.3&96.2&10.3 &20.5&93.5&40.5& 37.9 \\
MELO     && 94.9 &59.5& 64.8&92.4&17.4 &15.2&94.1 &40.7& 43.6 \\
WISE     && 95.8&78.4 & 76.2&82.3&47.4 &20.1&95.4 &71.2& 39.8 \\
\bottomrule
\end{tabular}%
\normalsize
\end{table}

\paragraph{Results on 2000 and 5000 sequential edits} We additionally report the editing performance under 2000 and 5000 sequential edits (batch size=1). For CounterFact, since the original fact $o_{\text{ori}}$ is provided, we additionally report probability-comparison-based metrics, where an edit is considered successful if $P(o_{\text{tar}} \mid x) > P(o_{\text{ori}} \mid x)$, following prior work~\citep{mengmass,fangalphaedit}. The corresponding metrics are denoted as \textbf{Eff*.}, \textbf{Gen*.}, and \textbf{Spe*.}.
As shown in Table~\ref{tab:results_2000} and Table~\ref{tab:editing_5000}, our proposed GEdit consistently achieves the best overall editing performance across all three backbone models (GPT2-XL, GPT-J, and LLaMA3) and both datasets (CounterFact and ZsRE).

Under 2,000 sequential edits, GEdit attains the highest or near-highest scores in Efficacy (Eff.), Generalization (Gen.), and Locality (Loc.) in almost every setting. For example, on GPT2-XL / CounterFact, GEdit achieves an \textit{Eff.} of 99.6 (tied with AlphaEdit) and a \textit{Gen.} of 94.0 (also tied best), while maintaining a \textit{Loc.} of 65.6. On LLaMA3, where baseline methods like EMMET and AlphaEdit suffer from catastrophic failure (\textit{Eff.} and \textit{Gen.} drop to 0 or 50\%), GEdit remains robust with \textit{Eff.} = 99.6, \textit{Gen.} = 93.0, and \textit{Loc.} = 78.1 on CounterFact.

The advantage of GEdit becomes even more pronounced under 5,000 sequential edits. While other methods degrade significantly --- for instance, on LLaMA3 / CounterFact, HSE shows a clear trend of failure, with SST dropping to around 50\%; on LLaMA3 / ZsRE, AlphaEdit also exhibits substantial degradation in performance --- GEdit maintains strong and stable results: \textit{Eff.} = 99.3, \textit{Gen.} = 92.3, \textit{Spe.} = 66.5 on CounterFact. Similarly, on GPT-J / ZsRE, GEdit achieves the highest scores among all methods across all three metrics (\textit{Eff.} = 99.6, \textit{Gen.} = 96.6, \textit{Spe.} = 27.3).

\paragraph{Results under ROME's evaluation setting}
Following ROME's evaluation protocol, we perform 10,000 independent edits. For each edit, we apply the edit to a fresh copy of the model, evaluate immediately, and then restore the model to its initial state before proceeding to the next edit. Performance metrics are then averaged across all edits. This setup tests whether a method can perform precise edits—achieving high efficacy while maintaining high locality.

Under this setting, all methods typically achieve high locality (\textit{Loc.}). Therefore, we adopt a more fine-grained metric, \textbf{CP.} (change in probability), defined as $\text{CP} = p(o \mid \text{original model}) - p(o \mid \text{edited model})$, where $o$ is the ground-truth answer for the locality test. A smaller CP indicates better preservation of unrelated knowledge.

As shown in Table~\ref{tab:precise_edit_results}, while all methods report high \textit{Loc.}, their \textit{CP.} scores differ significantly. Our method, \textsc{GEdit}, achieves the smallest \textit{CP.}, indicating superior locality, while also attaining the highest efficacy (\textit{Eff.}). This improvement stems from our adaptive regularization mechanism, which relaxes constraints on hard edits to ensure successful updates and tightens them on easy edits to better preserve pretrained knowledge.

\paragraph{Comparison with more baselines}
Beyond parameter-modifying methods, we further compare our approach against baselines from other paradigms, including memory-based and adapter-based editors. Specifically, we select SERAC~\citep{mitchell2022memory}, GRACE~\citep{hartvigsen2023aging}, MELO~\citep{yu2024melo}, and WISE~\citep{wang2024wise} as representative baselines. These methods share a common design principle: they employ external memory modules to store (or learn) edited facts. During inference, they retrieve and apply a stored fact based on the similarity between the input query and memorized entries.

Table~\ref{tab:editing_2000_memo} summarizes the performance of these editing methods under 2,000 sequential edits on the CounterFact and ZsRE benchmarks. Among the baselines, GRACE achieves the strongest locality—i.e., it minimally affects model predictions on unrelated inputs—yet suffers from poor generalization to paraphrased queries. This limitation stems from its memory-based design: GRACE stores key-value (KV) pairs for edited facts and, during inference, retrieves and replaces the corresponding forward activations whenever a match is detected. Although it avoids computing an explicit weight update, the retrieval and dynamic hidden activations substitution process introduces non-negligible overhead at inference time. MELO, like GRACE, is based on query-key matching, but instead of storing raw activations, it dynamically selects a learned LoRA adapter at inference time to apply the edit. In contrast, SERAC suffers from catastrophic forgetting and is therefore unsuitable for sequential editing. Both SERAC and WISE rely on trainable external modules—such as adapters or memory networks—that require additional training data for optimization. This not only increases the editing cost but also introduces computational latency during inference due to the extra forward pass through the auxiliary module.

Figure~\ref{fig:inference_time} reports the average inference time (in seconds) per input prompt across different backbone models. Our method consistently achieves the lowest latency on all three architectures, being up to 2.5$\times$ faster than SERAC on LLaMA3 (0.32\,s vs. 0.81\,s) and 4.6$\times$ faster on GPT2-XL (0.07\,s vs. 0.32\,s). This efficiency advantage stems from the parameter-modifying nature of our methods.

\newpage
\section*{NeurIPS Paper Checklist}

The checklist is designed to encourage best practices for responsible machine learning research, addressing issues of reproducibility, transparency, research ethics, and societal impact. Do not remove the checklist: {\bf The papers not including the checklist will be desk rejected.} The checklist should follow the references and follow the (optional) supplemental material.  The checklist does NOT count towards the page
limit. 

Please read the checklist guidelines carefully for information on how to answer these questions. For each question in the checklist:
\begin{itemize}
    \item You should answer \answerYes{}, \answerNo{}, or \answerNA{}.
    \item \answerNA{} means either that the question is Not Applicable for that particular paper or the relevant information is Not Available.
    \item Please provide a short (1--2 sentence) justification right after your answer (even for \answerNA). 
\end{itemize}

{\bf The checklist answers are an integral part of your paper submission.} They are visible to the reviewers, area chairs, senior area chairs, and ethics reviewers. You will also be asked to include it (after eventual revisions) with the final version of your paper, and its final version will be published with the paper.

The reviewers of your paper will be asked to use the checklist as one of the factors in their evaluation. While \answerYes{} is generally preferable to \answerNo{}, it is perfectly acceptable to answer \answerNo{} provided a proper justification is given (e.g., error bars are not reported because it would be too computationally expensive'' or ``we were unable to find the license for the dataset we used''). In general, answering \answerNo{} or \answerNA{} is not grounds for rejection. While the questions are phrased in a binary way, we acknowledge that the true answer is often more nuanced, so please just use your best judgment and write a justification to elaborate. All supporting evidence can appear either in the main paper or the supplemental material, provided in appendix. If you answer \answerYes{} to a question, in the justification please point to the section(s) where related material for the question can be found.

IMPORTANT, please:
\begin{itemize}
    \item {\bf Delete this instruction block, but keep the section heading ``NeurIPS Paper Checklist"},
    \item  {\bf Keep the checklist subsection headings, questions/answers and guidelines below.}
    \item {\bf Do not modify the questions and only use the provided macros for your answers}.
\end{itemize}


\begin{enumerate}

\item {\bf Claims}
    \item[] Question: Do the main claims made in the abstract and introduction accurately reflect the paper's contributions and scope?
    \item[] Answer: \answerYes{} 
    \item[] Justification: The abstract and introduction clearly state the main claims of the paper.

\item {\bf Limitations}
    \item[] Question: Does the paper discuss the limitations of the work performed by the authors?
    \item[] Answer: \answerYes{} 
    \item[] Justification: They have been discussed in Section \ref{sec:conclusion}.

\item {\bf Theory assumptions and proofs}
    \item[] Question: For each theoretical result, does the paper provide the full set of assumptions and a complete (and correct) proof?
    \item[] Answer: \answerYes{} 
    \item[] Justification: They have been provided in the Appendix.

    \item {\bf Experimental result reproducibility}
    \item[] Question: Does the paper fully disclose all the information needed to reproduce the main experimental results of the paper to the extent that it affects the main claims and/or conclusions of the paper (regardless of whether the code and data are provided or not)?
    \item[] Answer: \answerYes{} 
    \item[] Justification: The hyperparameter setting is clearly stated in the Section \ref{sec:setup}. 

\item {\bf Open access to data and code}
    \item[] Question: Does the paper provide open access to the data and code, with sufficient instructions to faithfully reproduce the main experimental results, as described in supplemental material?
    \item[] Answer: \answerYes{} 
    \item[] Justification: The data is publicly available and the anonymous link to the code is provided in the Appendix. 

\item {\bf Experimental setting/details}
    \item[] Question: Does the paper specify all the training and test details (e.g., data splits, hyperparameters, how they were chosen, type of optimizer) necessary to understand the results?
    \item[] Answer: \answerYes{} 
    \item[] Justification: They are introduced in Section \ref{sec:setup}.

\item {\bf Experiment statistical significance}
    \item[] Question: Does the paper report error bars suitably and correctly defined or other appropriate information about the statistical significance of the experiments?
    \item[] Answer: \answerNo{} 
    \item[] Justification: The error bar is not important in our research field as the randomness is not significant.

\item {\bf Experiments compute resources}
    \item[] Question: For each experiment, does the paper provide sufficient information on the computer resources (type of compute workers, memory, time of execution) needed to reproduce the experiments?
    \item[] Answer: \answerYes{} 
    \item[] Justification: See Appendix.
    
\item {\bf Code of ethics}
    \item[] Question: Does the research conducted in the paper conform, in every respect, with the NeurIPS Code of Ethics \url{https://neurips.cc/public/EthicsGuidelines}?
    \item[] Answer: \answerYes{} 
    \item[] Justification: The research conforms to the NeurIPS Code of Ethics. 

\item {\bf Broader impacts}
    \item[] Question: Does the paper discuss both potential positive societal impacts and negative societal impacts of the work performed?
    \item[] Answer: \answerNA{} 
    \item[] Justification: This is foundational methodology research with no direct path to specific societal applications or harms.
    
\item {\bf Safeguards}
    \item[] Question: Does the paper describe safeguards that have been put in place for responsible release of data or models that have a high risk for misuse (e.g., pre-trained language models, image generators, or scraped datasets)?
    \item[] Answer: \answerNA{} 
    \item[] Justification: The paper does not release high-risk models (e.g., pretrained language models, image generators) or scraped datasets, so no safeguards are required.

\item {\bf Licenses for existing assets}
    \item[] Question: Are the creators or original owners of assets (e.g., code, data, models), used in the paper, properly credited and are the license and terms of use explicitly mentioned and properly respected?
    \item[] Answer: \answerYes{} 
    \item[] Justification: All existing assets used (e.g., datasets, code libraries) are properly cited in Section \ref{sec:setup} and Appendix.

\item {\bf New assets}
    \item[] Question: Are new assets introduced in the paper well documented and is the documentation provided alongside the assets?
    \item[] Answer: \answerNA{} 
    \item[] Justification: This paper does not introduce new assets.

\item {\bf Crowdsourcing and research with human subjects}
    \item[] Question: For crowdsourcing experiments and research with human subjects, does the paper include the full text of instructions given to participants and screenshots, if applicable, as well as details about compensation (if any)? 
    \item[] Answer: \answerNA{} 
    \item[] Justification: This paper does not involve crowdsourcing or research with human subjects.

\item {\bf Institutional review board (IRB) approvals or equivalent for research with human subjects}
    \item[] Question: Does the paper describe potential risks incurred by study participants, whether such risks were disclosed to the subjects, and whether Institutional Review Board (IRB) approvals (or an equivalent approval/review based on the requirements of your country or institution) were obtained?
    \item[] Answer: \answerNA{} 
    \item[] Justification: No human subjects research was conducted, so IRB approval is not applicable.

\item {\bf Declaration of LLM usage}
    \item[] Question: Does the paper describe the usage of LLMs if it is an important, original, or non-standard component of the core methods in this research? Note that if the LLM is used only for writing, editing, or formatting purposes and does \emph{not} impact the core methodology, scientific rigor, or originality of the research, declaration is not required.
    \item[] Answer: \answerNA{} 
    \item[] Justification: LLMs were not used as an important, original, or non-standard component of the core methodology. Any LLM usage was limited to writing/editing assistance and does not affect scientific rigor or originality.

\end{enumerate}

\end{document}